\documentclass{article}
\usepackage{geometry}
\usepackage{amsmath, amssymb, amsthm}
\usepackage{graphicx}
\usepackage{hyperref}
\usepackage{natbib}
\usepackage{booktabs}

\theoremstyle{plain}
\newtheorem{theorem}{Theorem}[section]

\newtheorem{lemma}[theorem]{Lemma}

\theoremstyle{definition}
\newtheorem{definition}[theorem]{Definition}
\newtheorem{assumption}[theorem]{Assumption}
\theoremstyle{remark}
\newtheorem{remark}[theorem]{Remark}

\title{Towards Stable Machine Learning Model Retraining\\via Slowly Varying Sequences}

\author{
    Dimitris Bertsimas$^{1}$, Vassilis Digalakis Jr$^{2}$, Yu Ma$^{1}$, Phevos Paschalidis$^{3}$ \\
    $^{1}$Massachusetts Institute of Technology, Cambridge, US \\
    $^{2}$HEC Paris Business School, Jouy-en-Josas, France \\
    $^{3}$Harvard University, Cambridge, US 
}

\date{}

\begin{document}

\maketitle

\begin{abstract}
    We consider the problem of retraining machine learning (ML) models when new batches of data become available. Existing approaches greedily optimize for predictive power independently at each batch, without considering the stability of the model's structure or analytical insights across retraining iterations. We propose a model-agnostic framework for finding sequences of models that are stable across retraining iterations. We develop a mixed-integer optimization formulation that is guaranteed to recover Pareto optimal models (in terms of the predictive power-stability trade-off) with good generalization properties, as well as an efficient polynomial-time algorithm that performs well in practice. We focus on retaining consistent analytical insights --- which is important to model interpretability, ease of implementation, and fostering trust with users --- by using custom-defined distance metrics that can be directly incorporated into the optimization problem. We evaluate our framework across models (regression, decision trees, boosted trees, and neural networks) and application domains (healthcare, vision, and language), including deployment in a production pipeline at a major US hospital. We find that, on average, a 2\% reduction in predictive power leads to a 30\% improvement in stability.
\end{abstract}

\section{Introduction}
Machine learning models deployed in real-world production pipelines often require retraining to incorporate new data as it becomes available \cite{gama2014survey, wu2020deltagrad}. Retraining involves updating the model periodically to ensure its predictions remain accurate and aligned with the evolving data. However, in high-stakes domains such as healthcare, this process presents challenges. Frequent updates can lead to significant changes in the model's structure and behavior, undermining trust among expert users \cite{glikson2020human} and complicating the model's integration within operational pipelines \cite{baier2019challenges}. 

In sensitive applications, trust in AI-driven decision support systems is critical and is built over time. Physicians, policymakers, and other domain experts rely not only on a model’s accuracy but also on its consistency and alignment with domain knowledge \cite{dietvorst2018overcoming, Orfanoudaki2022}. Retraining that leads to unpredictable shifts in the model's structure or analytical insights can cause skepticism, hesitation in adoption, or even rejection of the model \cite{longoni2019resistance, babic2021beware}. Moreover, unstable retraining processes may result in legal or operational complications due to errors in critical decisions \cite{bertsimas2021algorithmic}.

Stability \cite{turney1995bias, bousquet2002stability} plays a pivotal role in overcoming these challenges. Stability in model structures ensures that changes between retraining iterations are smooth and interpretable, preserving trust in the system while allowing users to incorporate new insights without losing historical context. This is particularly important for high-stakes applications where consistent analytical insights are as valuable as predictive performance.


\subsection{Contributions}
We propose a framework to explicitly enforce stability in model retraining. Unlike existing methods, which focus primarily on maximizing predictive performance, our approach balances predictive power and structural stability. Our contributions are as follows:
\begin{itemize}
    \item \textbf{Framework for Retraining:} We propose a general, model-agnostic framework for retraining ML models that explicitly enforces stability across retraining iterations. The framework is also data-agnostic:
    \begin{itemize}
        \item If the data distribution remains stable over time, our framework prioritizes sequences of models that exhibit minimal structural changes, maintaining consistent insights.
        \item If the data distribution shifts significantly, our framework naturally favors models that better align with the new data, while still minimizing abrupt structural changes that undermine trust.
    \end{itemize}

    \item \textbf{Theoretical Guarantees:} We provide rigorous guarantees, showing that our framework produces solutions that are Pareto optimal or weakly Pareto optimal with respect to predictive performance and stability. We also establish a generalization bound for our methodology.

    \item \textbf{Algorithm for Practical Implementation:} Leveraging a mixed-integer optimization (MIO) formulation, we develop an efficient heuristic algorithm to make our framework practical for real-world applications. The algorithm accommodates a variety of model types, including logistic regression, classification trees, boosted trees, and neural networks.

    \item \textbf{Numerical Validation:} We validate our methodology across three real-world case studies, demonstrating its ability to improve stability with a small, controllable trade-off in predictive performance. This includes deployment in a production pipeline at a large US hospital, where our approach ensures stable mortality prediction models in a high-stakes healthcare setting.
\end{itemize}



\subsection{Related Work}

\textbf{Batch vs. Online Learning.}
Retraining ML models is a common requirement in production pipelines to incorporate new data and maintain predictive performance \cite{gama2014survey, wu2020deltagrad}. Depending on the data update frequency, the ML model retraining problem can be viewed either as a batch or an online learning problem. Batch learning involves periodically retraining models from scratch using the full dataset, whereas online learning incrementally updates models as new data becomes available \cite{Bisong2019, hoi2021online}. Online learning methods are computationally efficient but typically rely on partial data, resulting in greedier solutions that may not generalize as well. Recent works in online learning and adaptive learning \cite{gama2014survey} highlight their utility in dynamic environments, but these approaches often ignore structural consistency and require frequent updates that can disrupt production pipelines. In contrast, batch learning, by leveraging the entire dataset for retraining, ensures globally optimal solutions and smoother updates, making it better suited for high-stakes domains like healthcare, where interpretability, trust, and seamless integration with existing operational pipelines are critical.

\textbf{Model Retraining Challenges: Distribution Shifts and Costs.} 
Retraining ML models often requires addressing challenges such as detecting distribution shifts, managing retraining costs, and ensuring smooth transitions. Existing works focus on detecting shifts \cite{bifet2007learning, pesaranghader2016fast} or adapting models to mitigate their effects \cite{schwinn2022improving}; as shown by \cite{kabra2024limitations}, in settings where the deployed model influences the data distribution, naive retraining strategies can largely fail. Several works also consider retraining costs: \cite{mahadevan2023cost} propose an algorithm that balances retraining costs with performance deterioration by approximating a ``staleness'' cost, whereas \cite{vzliobaite2015towards} propose a return-on-investment framework for assessing retraining needs. However, these approaches primarily focus on deciding when to retrain rather than how to ensure consistency across retraining iterations, possibly resulting in significant changes to model behavior. Our work bridges these gaps by explicitly incorporating structural stability into the retraining process, enabling smooth transitions even under significant distribution shifts.

\textbf{Stable and Interpretable Models.}
The importance of stability for trust and interpretability has been recognized since early works \cite{turney1995bias, breiman1996heuristics}. Recent methods address stability either by training interpretable models with inherent structural consistency \cite{Dwyer_2007, bargagli2020causal, stable_tree, bertsimas2024slowly} or by ensuring stable feature interactions in black-box models \cite{Basu2018, wang2024stability}. 
Beyond classical ML models, \cite{zhang2020retrain} propose a retraining approach for recommender systems that, upon the arrival of new data, transfers past knowledge to the new model: while their method prioritizes efficiency and accuracy, it enforces an indirect form of stability by transferring learned experiences. These approaches are limited to specific model types or focus solely on feature importance without considering the stability of entire model sequences. Our work complements this by explicitly prioritizing structural stability alongside predictive performance, offering a unified framework that generalizes across model types and applications.

\section{A Framework for ML Model Retraining} \label{sec:meth}

\subsection{Problem Setting} \label{sec:problem-statement}
Consider an observational dataset accumulated over time composed of triplets \( (\boldsymbol{x_i}, y_i, t_i)_{i = 1}^N \), where \( \boldsymbol{x_i} \in \mathcal{X} \subseteq \mathbb{R}^p \) represents the feature vector, \( y_i \in \mathcal{Y} \subseteq \mathbb{R} \) is the target variable, and \( t_i \in \mathbb{R} \) is the time at which data point $i$ was obtained. The dataset is divided into a series of batches at times \( T_1 < T_2 < \ldots < T_B \). The dataset available at time \( T_b,\ b \in [B] := \{1,\ldots,B\}, \) is defined as \( D_b = \{ (x_i, y_i) : t_i \leq T_b, i \in [N] \}. \)
Thus, we have that \(D_1 \subseteq D_2 \subseteq \ldots \subseteq D_B\) and $\boldsymbol{D}= (D_1, \dots, D_B)$. We assume that $B\geq2$ and that, at model training time (denoted by \(t\)), all \(B\) batches are available, i.e., \(t \geq T_B\). In Section \ref{sec:meth:adap_retrain}, we discuss how to adaptively retrain the model using batches obtained after time \(t\). The goal is to find models \( \boldsymbol{f} := (f_1, f_2, \ldots, f_B) \in \mathcal{F}\times...\times\mathcal{F} := \mathcal{F}^B\) with \(f_b: \mathcal{X} \mapsto \mathcal{Y} \) for \(b\in [B]\) such that the following two objectives are satisfied: 

\textbf{Predictive power:} The first objective is to ensure that each model \( f_b \) in the sequence has small expected loss in terms of prediction quality. We denote by \( \mathbb{P}(D_b) \) the distribution of the training data in \(D_b\) and $\ell: \mathcal{Y} \times \mathcal{Y} \mapsto \mathbb{R}_{\geq0}$ the bounded loss function measuring, e.g., regression or classification accuracy; concretely, $\ell(y,f_b(\boldsymbol{x}))$ quantifies the quality of model $f_b$'s prediction on input $\boldsymbol{x}$ for $y$. Then, we seek for a sequence whereby each model (individually) achieves low expected loss $\mathcal{L}_P(f_b,D_b) = \mathbb{E}_{(\boldsymbol x, y) \sim \mathbb{P}(D_b)}[\ell(y, f_b(\boldsymbol x))].$ The overall loss of the sequence is then denoted by $\mathcal{L}_P(\boldsymbol{f},\boldsymbol{D}) = \sum_{b=1}^{B} \mathcal{L}_P(f_b,D_b).$ In practice, we approximate the expected loss using the empirical loss $\hat{\mathcal{L}}_P(f_b,D_b) = \frac{1}{N_b} \sum_{i: t_i \leq T_b}[\ell(y_i, f_b(\boldsymbol x_i))]$. The next assumption bounds the approximation quality as function of the sample size, the function class complexity, and the desired confidence: 
\begin{assumption} \label{assume:generalization}
    The predictive power objective satisfies a generalization bound of the form: {\small $\forall \delta \in (0,1)$, $\mathbb{P}\left( \forall f \in \mathcal{F}: |\mathcal{L}_P(f,D)-\hat{\mathcal{L}}_P(f,D)| \leq C(N, \mathcal{F}, \delta) \right) \geq 1-\delta$.}
\end{assumption}
        
\textbf{Stability:} The second objective is to ensure that consecutive models do not vary significantly, maintaining stability over time. Formally, we define a distance metric \( d: \mathcal{F} \times \mathcal{F} \mapsto \mathbb{R}_{\geq0} \) so that \(d(f_i, f_{j}) \) measures the change between models \(i\) and \(j\). The distance metric can be based on the models' parameters or structure (e.g., for linear regressions $\boldsymbol{\beta}_i,\boldsymbol{\beta}_j$, we can simply take $d(\boldsymbol{\beta}_i,\boldsymbol{\beta}_j)=\|\boldsymbol{\beta}_i - \boldsymbol{\beta}_j\|^2$), analytical insights (e.g., feature importances), or even predictions; we discuss the choice of distance metric in detail in Section \ref{sec:distance}. Then, for each pair of consecutive models, we minimize the stability loss $\mathcal{L}_S(\boldsymbol{f}) = \sum_{b=1}^{B-1} d(f_b, f_{b+1})$.
\begin{remark} \label{observe:generalization}
    The stability loss is data-independent, i.e., $\mathcal{L}_S(\boldsymbol{f},\boldsymbol D) = \mathcal{L}_S(\boldsymbol{f})$ for all data instances $\boldsymbol D$, so the stability   objective generalizes trivially, i.e., $\hat{\mathcal{L}}_S(\boldsymbol{f}) = \mathcal{L}_S(\boldsymbol{f})$.
\end{remark}

In practice, we are primarily interested in model \(f_B\) as this model has been trained on the full training set and is the one that will be used in production. Nevertheless, we choose to train the full sequence \(f_1, \ldots, f_B\) of \(B\) models since, by doing so, \textbf{(i)} we favor models that change smoothly upon the addition of new batches of data, and \textbf{(ii)} we are able to observe changing patterns, which is crucial for interpretability. Our approach thus relies on imposing \emph{stability by process}.

\subsection{General Formulation}
The objectives stated in Section \ref{sec:problem-statement} naturally give rise to a multi-objective optimization problem \cite{miettinen1999nonlinear}: we search for a sequence $\boldsymbol{f}$ that ``performs well'' in terms of both $\hat{\mathcal{L}}_P$ and $\hat{\mathcal{L}}_S$. This is commonly achieved by minimizing (i) a weighted combination of the objectives, i.e., $\lambda \hat{\mathcal{L}}_P + (1-\lambda)\hat{\mathcal{L}}_S$, where the weight $\lambda \in [0,1]$ is selected based on the learner's preferences (assuming these are known), or (ii) a preference-agnostic, min-max type formulation \cite{cortes2020agnostic}. In our setting, we often want to explicitly control ``the price of stability'' with respect to predictive power. This is particularly relevant in healthcare applications, where the accuracy requirements for using a model in production are stringent. Therefore, we opt for an $\epsilon$-constrained formulation: given a suboptimality tolerance parameter $\alpha \geq 0$, the predictive error of each model \( f_b \) should not be worse than the best model for batch \(D_b\) by more than a multiplicative factor of $\alpha$. Recalling our assumption that, at training time, we have already received \(B\geq2\) batches of data (if this is not the case, we artificially split the training data into \(B\) batches), we obtain the following formulation:
\begin{equation} \label{eqn:formulation-constrained}
\begin{split}
    \min_{f_1, \ldots, f_B \in \mathcal{F}^B} & \quad  \hat{\mathcal{L}}_S(\boldsymbol{f}) \ := \ \sum_{b=1}^{B-1} d(f_b, f_{b+1}) \\
    \text{s.t.} & \quad \hat{\mathcal{L}}_P(f_b, D_b) \leq (1 + \alpha) \cdot  \\
    & \quad \cdot \min_{f\in \mathcal{F}}\hat{\mathcal{L}}_P(f, D_b) := \epsilon_b \quad \forall b \in [B].
\end{split}
\end{equation}

\subsection{Pareto Optimality of Solutions}
To analyze the quality of solutions to Problem \eqref{eqn:formulation-constrained}, we adapt the definition of Pareto optimality:
\begin{definition}
    A sequence $\boldsymbol{f} \in \mathcal{F}^B$ is \textbf{weakly Pareto optimal} (WPO) if there is no other sequence $\boldsymbol{f'}$ such that $\hat{\mathcal{L}}_P(f'_b, D_b) <\hat{\mathcal{L}}_P(f_b,D_b)\ \forall b \in [B]$ and $\hat{\mathcal{L}}_S(\boldsymbol{f'}) < \hat{\mathcal{L}}_S(\boldsymbol{f})$. $\boldsymbol{f}$ is \textbf{Pareto optimal }(PO) if the inequality is strict for at least one objective and non-strict for the remaining ones.
\end{definition}
The next results, which we prove in Appendix \ref{appx:technical-proofs}, show that any solution to Problem \eqref{eqn:formulation-constrained} is WPO, whereas a unique solution is PO:
\begin{theorem} \label{theo:wpo}
    A sequence $\boldsymbol{f^*}$ obtained by solving Problem \eqref{eqn:formulation-constrained} is WPO.
\end{theorem}
\begin{theorem} \label{theo:po}
    A sequence $\boldsymbol{f^*}$ is PO if it is a unique solution to Problem \eqref{eqn:formulation-constrained} for any given $\boldsymbol{\epsilon}$.
\end{theorem}

In light of Theorems \ref{theo:wpo} and \ref{theo:po}, the proposed methodology is guaranteed to produce PO sequences for models that correspond to strongly convex optimization problems (e.g., $\ell_2$-regularized logistic regression) and WPO sequences for all models. As we show next (proof in Appendix \ref{appx:technical-proofs}), to verify PO for non-strongly convex optimization problems, it suffices to solve a series of ``complementary problems,'' in each of which the predictive power loss for one batch is minimized subject to appropriately chosen constraints on the remaining predictive power losses and the stability loss:
\begin{theorem} \label{theo:pov}
    A sequence $\boldsymbol{f^*}$ is PO if and only if it solves both Problem \eqref{eqn:formulation-constrained} with $\epsilon_b = \hat{\mathcal{L}}_P(f^*_b, D_b) := \epsilon^*_b \ \forall b \in [B]$ and the following problem for all $b'\in [B]$: 
    \begin{equation} \label{eqn:formulation-complementary}
    \begin{split}
        \min_{\boldsymbol{f}} & \quad \hat{\mathcal{L}}_P(f_{b'}, D_{b'}) \\
        \text{s.t.} & \quad \hat{\mathcal{L}}_P(f_b, D_b) \leq \epsilon^*_b \quad \forall b \not= b' \\
        & \quad \hat{\mathcal{L}}_S(\boldsymbol{f}) \leq \hat{\mathcal{L}}_S(\boldsymbol{f}^*).
    \end{split}
    \end{equation}
\end{theorem}
In all, we can recover the entire (empirical) Pareto frontier by varying the value of the accuracy tolerance parameter in Problem \eqref{eqn:formulation-constrained}, and possibly solving a series of complementary problems for verification. 

\subsection{Pareto Excess Risk Bound} \label{ssec:pareto-excess-risk}
The next result, which follows by direct application of Theorem 4 from \cite{sukenik2022generalization}, guarantees that, by finding the empirical (in-sample) Pareto frontier, we approximately recover the true (out-of-sample) Pareto frontier with respect to all objectives:
\begin{lemma} \label{lem:generalization}
    Under Assumption \ref{assume:generalization} and Remark \ref{observe:generalization}, for any $\delta \in (0,1)$, with probability at least $1-\delta$, for all PO sequences $\boldsymbol{f^*}$ that solve Problem \eqref{eqn:formulation-constrained} using the expected losses $\mathcal{L}_P$ and $\mathcal{L}_S$, there exists a PO sequence $\boldsymbol{\hat{f^*}}$ that solves Problem \eqref{eqn:formulation-constrained} using the empirical losses $\hat{\mathcal{L}}_P$ and $\hat{\mathcal{L}}_S$, such that the following hold:
    $ \mathcal{L}_P(\hat{f^*_b},D_b) \leq \mathcal{L}_P(f^*_b,D_b) + 2 C(N,\mathcal{F},\frac{\delta}{B}),\ \forall
    b \in [B]$ and $\mathcal{L}_S(\boldsymbol{\hat{f^*}}) = \mathcal{L}_S(\boldsymbol{f^*})$.
\end{lemma}

For example, for linear regression, a standard finite-sample result states that, with probability at least $1 - \delta$, the excess risk of the empirical risk minimizer of batch $b$, $\hat{f}_b$, satisfies:
\begin{equation} \label{eq:linear-regression-hp-bound}
    \mathcal{L}_P(\hat{f}_b, D_b) - \mathcal{L}_P(f^*_b, D_b) \leq \frac{\sigma^2 p}{N_b} + \sigma \sqrt{\frac{4 p \log(1/\delta)}{N_b}}.
\end{equation}
This bound holds independently for each batch, ensuring that each $\hat{f}_b$ generalizes well to new data. In our setting, however, we seek to assess the Pareto optimality of the full sequence $(f_1, \dots, f_B)$. By applying Lemma \ref{lem:generalization}, we obtain the following multi-objective extension of \eqref{eq:linear-regression-hp-bound}:
{\small \begin{equation} \label{eq:multi-objective-bound}
    \mathcal{L}_P(\hat{f^*_b},D_b) - \mathcal{L}_P(f^*_b,D_b) \leq \frac{2\sigma^2 p}{N_b} + 2 \sigma \sqrt{\frac{4 p \log\frac{B}{\delta}}{N_b}}, \ \forall b \in [B].
\end{equation}}
This result guarantees that the entire sequence $(\hat{f}_1, \dots, \hat{f}_B)$ remains Pareto optimal with high probability. Ensuring generalization across the entire sequence results in a slightly larger excess risk bound due to the tighter confidence level $\frac{\delta}{B}$, growing logarithmically with the number of models $B$; at the same time, the increasing sample size $N_b$ guarantees that the empirical Pareto frontier still converges to the true Pareto frontier. We numerically study the Pareto excess risk bound for linear regression in Appendix \ref{appx:excess-risk-bound}.

\section{A Practical Algorithm for Model Retraining}

\subsection{Tractable Restricted Formulation} \label{sec:meth:tract}
Depending on the type of the model used, solving Problem \eqref{eqn:formulation-constrained} can be challenging: the number of possible functions $f_b$ may grow exponentially with the number of features. This is particularly true for models that are commonly used in high-stakes healthcare applications, including decision trees and variants thereof: considering $p$ binary features and a binary classification task, then there are $2^{2^p}$ different ways of labeling all instances, each corresponding to a different underlying boolean function that can be represented as a decision tree. Thus, we next develop a restricted formulation that can tractably accommodate any type of ML model.

For each batch \( b \in [B] \), we pre-train a finite set of candidate models, denoted as \( \tilde{\mathcal{F}}_b \). Then, the problem reduces to selecting one model from each set of pre-computed models to form a sequence. Let \( f_{b, j} \) be the \( j \)-th model in \( \tilde{\mathcal{F}}_b \), and introduce binary variables \( z_{b, j} \) such that \( z_{b, j} = 1 \) if model \( f_{b, j} \) is selected, and 0 otherwise. This yields the following MIO formulation:
{\small 
\begin{equation} \label{eqn:formulation-restricted}
\begin{split}
    \min_{\substack{f_{b, j} \in \tilde{\mathcal{F}}_b \\ z_{b, j} \in \{0, 1\}}} & \quad \sum_{b=1}^{B-1} \sum_{j=1}^{|\tilde{\mathcal{F}}_b|} \sum_{k=1}^{|\tilde{\mathcal{F}}_{b+1}|} d(f_{b, j}, f_{b+1, k}) \cdot z_{b, j} \cdot z_{b+1, k} \\
    \text{s.t.} & \quad \sum_{j=1}^{|\tilde{\mathcal{F}}_b|} \mathcal{L}_P(f_{b, j}, D_b) \cdot z_{b, j} \leq (1 + \alpha) \cdot \\
    & \qquad \cdot \min_{f \in \tilde{\mathcal{F}}_b} \mathcal{L}_P(f, D_b) \quad \forall b \in [B] \\
    & \quad \sum_{j=1}^{|\tilde{\mathcal{F}}_b|} z_{b, j} = 1 \quad \forall b \in [B].
\end{split}
\end{equation}
}

The following theorem allows us to solve the restricted problem of selecting a slowly varying sequence of models (Problem \eqref{eqn:formulation-restricted}) in polynomial time (proof in Appendix \ref{appx:technical-proofs}):
\begin{theorem}\label{theo:shortest-path-reduction} Problem \eqref{eqn:formulation-restricted} can be reduced to a shortest path problem in a directed graph.
\end{theorem}

\begin{figure}[h]
    \centering
    \includegraphics[width=\linewidth]{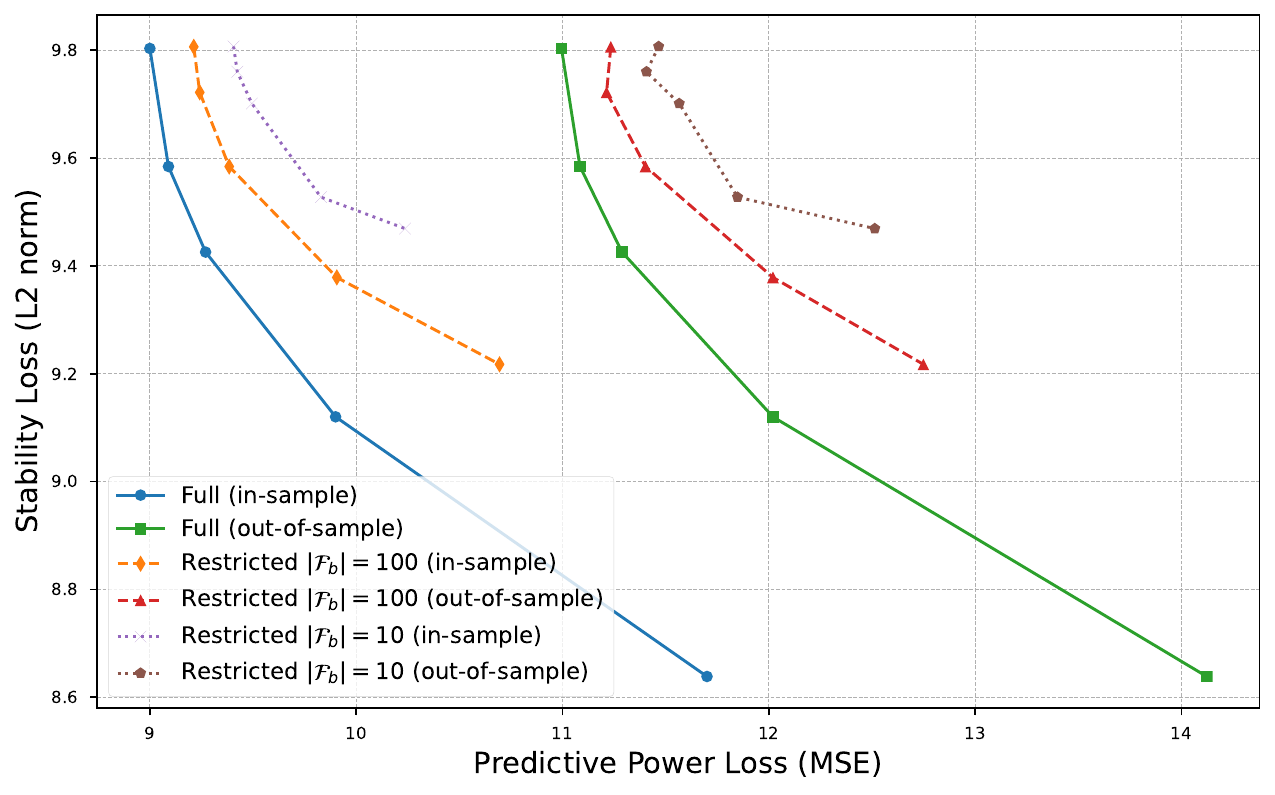} 
    \caption{Pareto frontier (in-sample and out-of-sample) between predictive power (aggregated across batches) and stability, obtained by solving Problem \eqref{eqn:formulation-constrained} (full) or Problem \eqref{eqn:formulation-restricted} (restricted for varying number of candidate models)}
    \label{fig:pareto}
\end{figure}

The restricted formulation (Problem \eqref{eqn:formulation-restricted}) is an approximation of the full formulation (Problem \eqref{eqn:formulation-constrained}). To assess the quality of the approximation and the impact of the size of $\mathcal{F}_b$ (number of candidate models per batch) on the restricted formulation, we train slowly varying sequences of linear regression models on synthetic data generated according to $y = \boldsymbol{x}^\top \boldsymbol{\beta}_{b} + \epsilon$, where, for each batch, the data $X\in\mathbb{R}^{n\times p} (\text{with } n=100, p=10)$ is drawn from $\boldsymbol{x} \sim \mathcal{N}(\boldsymbol{0},\Sigma), \epsilon \sim \mathcal{N}(0,\sigma)$ and the regression coefficients satisfy $|\boldsymbol \beta_b - \boldsymbol \beta_0| \leq 0.3$ for some $\boldsymbol \beta_0 \in \{-1,0,1\}$. Note that, thanks to the convexity of the linear regression optimization problem, by varying the value of $\epsilon_b$ in Problem \eqref{eqn:formulation-constrained}, we recover the full empirical Pareto frontier. Figure \ref{fig:pareto} presents the empirical (in-sample) Pareto frontier obtained by solving the full formulation (Problem \eqref{eqn:formulation-constrained}) as well as the restricted formulation (Problem \eqref{eqn:formulation-restricted}) for increasing number of candidate models per batch. We also plot the respective true (out-of-sample) Pareto frontiers---visualizing the generalization properties of the proposed approach.

We summarize the key findings from Figure \ref{fig:pareto}:
\begin{itemize}
    \item \textbf{Convexity of the Pareto Frontier:} The Pareto frontier is convex, meaning that a small increase in predictive power loss leads to a significant reduction in stability loss. This demonstrates the potential for controlled trade-offs between performance and stability. 
    
    Specifically, to assess the effect of the accuracy tolerance $\alpha$ on the stability gains, we examine the full model’s in-sample curve. We evaluate accuracy tolerances \( \alpha \in \{0, 0.01, 0.03, 0.1, 0.3\} \), meaning that the predictive power validation loss is constrained to be within \( \alpha \) of the best independent model for each batch. The corresponding gains in stability loss range between 2-12\%, confirming that moderate increases in predictive loss can substantially enhance stability.

    \item \textbf{Comparison of Full vs. Restricted Models:} The restricted model does not always recover PO solutions due to finite model sampling per batch, potentially missing the optimal sequence for certain accuracy tolerances. Consequently, the restricted model finds solutions that are Pareto-dominated by those of the full model. However, increasing \( |\mathcal{F}_b| \) improves the approximation, allowing the restricted model to better approach the full model’s solutions.
    \begin{itemize}
        \item For \textit{small accuracy tolerances} (\(\alpha \leq 0.03\)), which are the most relevant in practice, the restricted models are 1-3\% worse than the full model in both predictive power and stability.
        \item For \textit{higher accuracy tolerances}, the restricted model fails to recover solutions where its stability loss matches that of the full model. This limitation arises because the restricted model explores a finite subset of candidates per batch, making it harder to find sequences with minimal structural change. Achieving stability loss comparable to the full model would require impractically large candidate sets \( |\mathcal{F}_b| \), effectively ensuring that consecutive batches contain nearly identical models.
    \end{itemize}

    \item \textbf{Generalization Performance:} The in-sample vs. out-of-sample predictive power gap is around 20\%. Notably, the gap between the full and restricted models remains unchanged between in-sample and out-of-sample settings, validating the theoretical generalization guarantees. Most importantly, the observed difference between in-sample and out-of-sample losses empirically supports the quality of the empirical Pareto frontier as a reasonable approximation of the true Pareto frontier.
\end{itemize}

\subsection{Adaptive Model Retraining} \label{sec:meth:adap_retrain}
We now discuss how to apply this framework in the context of adaptive model retraining: when a new batch of data becomes available and we wish to update an existing slowly varying sequence to incorporate the new data. More specifically, we have already trained an existing slowly varying sequence $f_1, f_2, \dots, f_B$ using an observational dataset $D_B$. We are now given an additional batch of data, which we append to the existing dataset to create $D_{B+1}$. We want to obtain a model $f_{B+1}$ that remains close in distance to the current model, $f_B$.

To do so, we construct an independent slowly varying sequence $f'_{1}, f'_{2}, \dots, f'_{B+1}$ created on the updated dataset, $D_{B+1}$, and we deploy $f'_{B+1}$. Unlike the perhaps more obvious choice of greedily choosing a new model $f^{\mathrm{greedy}}_{B+1}$ trained on $D_{B+1}$ that is close to $f_{B}$, the proposed approach achieves stability \textit{by process} and generalizes to longer deployment horizons with multiple model updates and can also be used when re-training after receiving multiple data updates --- e.g., constructing $f'_{B+m}$ when the existing model is $f_{m}$. Next, we formalize the ``stability by process'' argument:
\begin{definition}
    The \textbf{intra-sequence stability} loss is defined as the sum of pairwise distances of adjacent models of one sequence, i.e., $\mathcal{L}_S(\boldsymbol{f}) = \sum_{b=1}^{B-1} d(f_b, f_{b+1})$.
\end{definition}
\begin{definition}
    The \textbf{inter-sequence stability} loss is defined as the distance between the final models of two sequences of lengths $B$ and $B'$ respectively, i.e., $\mathcal{L}_S(\boldsymbol{f}, \boldsymbol{f'}) = d(f_B, f'_{B'})$.    
\end{definition}
For the inter-sequence stability loss, we could alternatively compare all pairs of models in $\boldsymbol{f}, \boldsymbol{f'}$. However, such an approach (i) does not directly apply to sequences of different lengths and (ii) ignores the fact that we mostly care about the final model from each sequence (which is used in production). The next result promises that, under certain conditions, intra-sequence stability implies inter-sequence stability.
In particular, we consider the implications of training a sequence of models that minimizes the intra-sequence stability loss on the inter-sequence stability when a new batch of data is introduced, and the entire sequence is retrained.
We work under the additional assumption that the initial model is common in both sequences, i.e., $f_1 = f_1' = f_0$. This is a realistic assumption as, in practice, $f_0$ can be the model that was previously used in production. Further, this assumption can easily be incorporated as an extra constraint in the proposed optimization formulation.
\begin{theorem} \label{theo:intra-inter}
    Let $\boldsymbol{f}, \boldsymbol{f'}$ be two sequences of models trained using data $D_1, \dots, D_B$ and $D_1, \dots, D_B, D_{B+1}$, respectively, and a stability loss metric $d: \mathcal{F} \times \mathcal{F} \mapsto \mathbb{R}_{\geq 0}$.
    Assume (i) their intra-sequence stability losses satisfy $\mathcal{L}_S(\boldsymbol{f}) \leq \epsilon$ and $\mathcal{L}_S(\boldsymbol{f
    '}) \leq \epsilon'$, and (ii) the initial models satisfy $f_1 = f_1' = f_0$.
    Then, $\mathcal{L}_S(\boldsymbol{f}, \boldsymbol{f'}) \leq \epsilon + \epsilon'$.
\end{theorem}

\section{Measuring the Stability Loss} \label{sec:distance}

To measure the stability loss $\mathcal{L}_s(\boldsymbol{f})$ of a sequence $\boldsymbol{f}$, we need to compute the distance $d(f_i, f_j)$ between pairs of models $f_i$ and $f_j$. We consider two primary approaches: structural distances and analytical insight-based distances.

\subsection{Structural Distances}
This approach measures the distance by directly comparing the models' structures. Thus, structural distances are model dependent. We discuss structural distances for three model families: regression models, decision tree-based models, and neural networks. 

\textbf{Regression Models.} 
We compute the distance as $d(f_i, f_j) = \| \boldsymbol \beta_i - \boldsymbol \beta_j \|_2^2$, where $\boldsymbol \beta_i$ and $\boldsymbol \beta_j$ are the regressions coefficient vectors of $ f_i$ and $ f_j$, respectively. We focus on linear and logistic regression---though our approach directly applies to generalized linear models as well.

\textbf{Decision Tree-based Models}.
We use the decision tree distance metric proposed by \cite{stable_tree} that relies on optimally matching the paths in the two trees $\mathbb{T}_i$ and $\mathbb{T}_j$, denoted as $\mathcal{P}(\mathbb{T}_i)$ and $\mathcal{P}(\mathbb{T}_j)$. The distance computation is as follows: first, we calculate the pairwise distance between each path in the first tree and each path in the second tree $d(p, q)$; this pairwise distance takes into account the differences in feature ranges in the two paths as well as the class labels associated with them. Then, we determine the overall distance between two decision trees by finding the optimal matching of these paths, a process that can be computed in polynomial time as a variant of the bipartite matching problem. Formally, assuming without loss of generality that $|\mathcal{P}(\mathbb{T}_i)| \geq |\mathcal{P}(\mathbb{T}_j)|$, and given decision variables $x_{pq}=  \mathbb{I}$(path $p$ in $\mathbb{T}_i$ is matched with path $q$ in $\left.\mathbb{T}_j\right)$, the distance is defined as the following (polynomially solvable) linear optimization problem:
{\small
\begin{equation} \label{eq:distance}
\begin{aligned}
    & \min_{x_{pq} \in [0,1]} \ \sum_{p \in \mathcal{P}(\mathbb{T}_i)} \sum_{q \in \mathcal{P}(\mathbb{T}_j)} d(p,q) x_{pq} \qquad \text{s.t.}  \\
    & \quad \sum_{q \in \mathcal{P}(\mathbb{T}_j)} x_{pq} = 1, \forall p \in \mathcal{P}(\mathbb{T}_i),  
    \ \sum_{p \in \mathcal{P}(\mathbb{T}_i)} x_{pq} \geq 1, \forall q \in \mathcal{P}(\mathbb{T}_j).
\end{aligned}
\end{equation}

\textbf{Neural Networks.}  
To measure the structural difference between neural networks, we compare the singular values of their weight matrices. The intuition behind this approach, inspired by \cite{raghu2017svcca}, is that the singular values of a weight matrix determine how the network transforms input data, describing its ability to stretch, compress, or rotate different feature directions. For each pair of neural networks $f_i, f_j$, we compute the spectral distance by summing the differences in singular values across all corresponding weight matrices:
$d(f_i, f_j) = \sum_{l} \|\sigma(W_i^{(l)}) - \sigma(W_j^{(l)})\|_2,$
where $W_i^{(l)}$ and $W_j^{(l)}$ are the weight matrices of layer $l$ in networks $f_i$ and $f_j$ and $\sigma(W)$ denotes the ordered singular values of matrix $W$. This metric captures how similarly two networks process data at each layer: small spectral distance implies similar feature transformations, whereas large spectral distance suggests significant differences in how information is propagated.

For comparing neural networks with similar architectures but differing layer sizes, one approach is padding weight matrices with zeros to ensure dimensional consistency before computing singular values. However, when architectures differ significantly, designing a robust distance metric is challenging; we leave the development of such methods for future work.

\subsection{Analytical Insight-based Distances}
Feature importance is a widely used tool for assessing the analytical insights provided by a model, as it helps explain which features drive predictions. To quantify the stability of these insights across retraining iterations, we measure the distance between models by comparing their feature importance vectors. Given a model $f$, let $\boldsymbol{\phi}_f \in \mathbb{R}^p$ denote the importance assigned to each of the $p$ features. We compute the distance as  $d(f_i, f_j) = \|\boldsymbol{\phi}_{f_i} - \boldsymbol{\phi}_{f_j} \|_2^2.$

This approach is applicable to both interpretable and black-box models. E.g., for decision tree-based models, we leverage model-specific feature importance metrics such as Gini importance, which measures the contribution of a feature to reducing node impurity, and Gain, which quantifies improvements in classification accuracy from splits made by each feature \cite{breiman1984classification, pedregosa2011scikit}. More generally, for any model type---including black-box models---we can compute feature importance distances using SHAP (SHapley Additive exPlanations) \cite{shap}, which offer a standardized methodology to measure feature importance by calculating the average marginal contribution of each feature to the prediction. This provides a unified, model-agnostic way to compare analytical insights across model types.

\subsection{Comparing Distance Types}
In Appendix \ref{appx:distance-experiment}, we analyze the relationship between structural stability, analytical insights, and predictive consistency. We find that the proposed structural distances correlate with both analytical insight-based distances and differences in predictions, confirming that shifts in a model's learned structure translate into changes in feature attribution and output stability. Logistic regression and decision trees exhibit strong alignment between structural and analytical insight-based stability, whereas neural networks show greater structural variation with lower feature importance shifts, suggesting a more complex relationship between learned weights and interpretability. Across all models, we observe that low structural and analytical insight-based distances often imply similar predictions---but the reverse is not necessarily true; this suggests that the proposed stability loss metrics offer stronger guarantees compared to common prediction-based ones.

\section{Numerical Experiments} \label{s:experiments}

\subsection{Case Studies} \label{ss:setting}
We validate our methodology through three case studies across different domains: in \textbf{healthcare}, we deploy the proposed methodology in production in collaboration with a large US hospital; in \textbf{vision}, we analyze model stability in a facial recognition task using the Yearbook dataset; and in \textbf{language}, we perform dynamic news categorization using the HuffPost dataset \cite{yao2022wild}.

\textbf{Healthcare.} 
We collect daily patient data from a large hospital system in Connecticut. We extract patient features including demographics, patient status, clinical measurements, lab results, diagnoses, orders, procedures, and notes. Patient records are retrospectively identified from January 1st, 2018 to May 1st, 2022, overseeing 168,815 patients, and 865,954 patient-day records. We focus on the binary classification task of predicting patient mortality risk, defined as the likelihood of death or hospice care by hospital discharge. We evaluate model performance using the Area Under the Receiver Operating Characteristic Curve (AUC), which assesses the model's ability to differentiate between outcomes by comparing the true positive rate against the false positive rate at various thresholds.

\textit{Data preprocessing.} In the existing production pipeline, models are trained on a 6-month basis, although the retraining window is scheduled to be reduced to 3-4 months. There are no obvious data shifts for this case study. To simulate the process of receiving data updates, we divide the 52 months into multiple sub-intervals. We follow traditional ML training procedures by dividing the dataset into training, validation, and testing sets. Specifically, all patient records admitted from September 2021 to December 2022 (16 months) are allocated as the out-of-sample testing set.


\textbf{Vision.} Consisting of 37,921 American high school yearbook photos, the Yearbook dataset encompasses the evolving social norms, fashion styles and population demographics over the period of 1930-2013. Each photo is a \(32 \times 32 \times 1\) gray-scale image associated with a binary outcome of the individual's gender. 

\textit{Data preprocessing.} Similar to the approach for the healthcare dataset, we divide the full time sequence into training, validation, and testing given their time sequence. There are 84 years in total, and we consider every 7 years as a single time period. All photos from and including 2010 are allocated as the out-of-sample testing set. Instead of feeding the raw image as our input, we first process embeddings using open-source, pre-trained vision models that are finetuned on similar datasets. Specifically, we process the existing image to be compatibly shaped tensored to be passed to FaceNet, an existing Inception Residual Masking Network pretrained on VGGFace2 to classify facial identities. Each image is reduced down to an embedding of the shape 512, which are then used to fit the ML models of which we study stability tradeoffs on.

\textbf{Language.} The Huffpost dataset contains around 200,000 samples and aims to identify tags of news articles from their corresponding headlines, which are classified into 11 different categories (Black Voices, Business, Comedy, Crime, Entertainment, Impact, Queer Voices,
Science, Sports, Tech, Travel). 

\textit{Data preprocessing.} The dataset ranges from 2012 to 2018, where each year is considered as a single individual time period in the experiment. All news headlines of 2018 are allocated as out-of-sample testing set. Similar to the vision dataset, instead of directly using a large language model to study their stability, we first process each sample using a pretrained model to obtain their respective embeddings. In particular, we use the BAAI FlagEmbedding model to map the original text into lower-dimensional vectors of size 2014.

\subsection{Research Questions and Methodology}

\textbf{Research questions.} Our main experiment serves to illuminate the predictive power-stability trade-off that the slowly varying methodology (SVML) hopes to optimize. To assess the performance of our methodology with respect to this trade-off, we evaluate the predictive power and the intra-sequence stability of the generated sequences, and compare them against those of a greedily trained sequence. Furthermore, as secondary research questions, we perform the following robustness checks: \textbf{(i)} first, we analyze the stability of the adaptive model retraining strategy by looking at the inter-sequence stability of the generated sequences; \textbf{(ii)} second, we investigate the sensitivity of SVML to the update frequency (or interval length); \textbf{(iii)} third, we examine the robustness of SVML with respect to different distance measures and the consistency of the resulting analytical insights. 

\textbf{Experimental methodology.} We obtain experimental results as follows. Having selected an update frequency, the number of batches $B$ is fixed. For each batch $b \in B$, we define the available training data $D_b$ as the first $b$ intervals and reserve the $(b+1)$-th interval as a validation set. We generate a set of pre-computed candidate models, $\mathcal{F}_b$, for each batch, bootstrapping to ensure diversity in the candidate models. Then, for accuracy tolerances [0.1, 0.05, 0.02, 0.01], we generate a slowly varying sequence according to the methodology outlined in Section \ref{sec:meth:tract}. To benchmark the performance of these sequences, we also generate a ``greedy sequence'' constructed of greedily selected, best-performing models (evaluated on the validation set). We denote by $f_b$ the batch $b$ model chosen by the slowly varying sequence and denote by $\bar{f}_b$ the batch $b$ model chosen by the greedy sequence. All AUC results are reported on the 16-month testing set put aside at the start of model training.

\textbf{Experimental software.}  We implement all algorithms in \verb|Python| 3.7 and run all experiments on a standard Intel(R) Xeon(R) CPU E5-2690 @ 2.90GHz running CentOS release 7. 

\subsection{Results}

\paragraph{Trading-off Predictive Power and Stability.} We start with an in-depth analysis of the distance-accuracy trade-off implicit in the slowly varying methodology.  we compare the performance of our slowly varying machine learning (SVML) methodology against a greedy retraining approach across the three datasets (Yearbook, Healthcare, Huffpost) and four model families (logistic regression, decision trees, boosted trees, and neural networks). The results are summarized in Tables~\ref{tab:stability_comparison} and~\ref{tab:accuracy_comparison}. We include a detailed presentation of the results in Appendix \ref{appx:additional-results}.

Table~\ref{tab:stability_comparison} presents the percent improvement in stability loss $\mathcal{L}_S$ as a function of the accuracy tolerance $\alpha$. It quantifies the extent to which SVML reduces model instability across retraining iterations, demonstrating that greater stability is achieved as the accuracy threshold increases. In some cases, the same models were selected for different thresholds, which is expected when the thresholds were not large enough to shift the optimal solution. Table~\ref{tab:accuracy_comparison}, on the other hand, quantifies the percent deterioration in realized out-of-sample accuracy due to applying SVML instead of a greedy retraining approach. We make the following observations:
\begin{itemize}
    \item On average, across all datasets and model types, a $2\%$ reduction in (realized) predictive power results in a $30\%$ improvement in model stability.
    \item As the accuracy tolerance increases, stability gains become more pronounced, confirming that allowing small sacrifices in predictive power leads to smoother model updates.
    \item The degree of stability improvement varies by model type, with interpretable models (logistic regression, CART) and neural networks benefiting more from stability constraints than ensemble models like XGBoost (which are known to be more stable).
    \item The actual, realized deterioration in out-of-sample predictive power is always significantly smaller than the pre-selected accuracy tolerance, suggesting that stabilization often leads to improved generalization. (E.g., negative deterioration is due to the fact that the greedy methodology selects the best performing model using a \textit{validation set}, so, when evaluated on the test set, its performance could be worse that SVML.)
\end{itemize}
To summarize, \textbf{(i)} across all models and datasets, as we allow a higher accuracy tolerance, the improvement in stability consistently increases. This is expected, as relaxing the predictive power constraint enables SVML to prioritize selecting smoother model transitions. \textbf{(ii)} While all models benefit from the SVML approach, the stability gains are most pronounced for logistic regression, decision trees, and neural networks. These models exhibit inherently higher sensitivity to data perturbations, and thus benefit more from explicitly enforcing stability. \textbf{(iii)} The average results confirm that even for small sacrifices in predictive power (1-2\%), we can achieve meaningful reductions in model instability. This is important in real-world deployment settings, where sudden changes in model outputs can lead to user distrust or operational inefficiencies.

\begin{table}[h]
    \centering
    \resizebox{\linewidth}{!}{%
    \begin{tabular}{lcccc|cccc|cccc|cccc}
        \toprule
        \textbf{Dataset} & \multicolumn{4}{c|}{Yearbook} & \multicolumn{4}{c|}{Healthcare} & \multicolumn{4}{c|}{Huffpost} & \multicolumn{4}{c}{\textbf{Avg.}} \\
        \textbf{$\alpha$} & 0.01 & 0.02 & 0.05 & 0.1 
        & 0.01 & 0.02 & 0.05 & 0.1 
        & 0.01 & 0.02 & 0.05 & 0.1 
        & 0.01 & 0.02 & 0.05 & 0.1 \\
        \midrule
        LogReg & 23.76 & 23.76 & 23.76 & 23.76 & 46.67 & 60.86 & 69.84 & 69.84 & 8.74 & 8.74 & 8.74 & 8.74 
        & 26.39 & 31.12 & 34.11 & 34.11 \\
        CART & 15.50 & 26.16 & 26.91 & 26.91 & 23.86 & 25.02 & 29.67 & 29.67 & 22.36 & 26.65 & 30.8 & 30.8
        & 20.57 & 25.94 & 29.13 & 29.13 \\
        XGBoost & 11.09 & 11.09 & 11.09 & 11.09 & 23.43 & 25.03 & 25.03 & 25.03 & 7.17 & 7.17 & 7.17 & 7.17 
        & 13.90 & 14.43 & 14.43 & 14.43 \\
        MLP & 18.29 & 23.21 & 23.21 & 23.21 & 34.3 & 49.14 & 73.53 & 83.99 & 33.62 & 33.62 & 33.62 & 33.62 
        & 28.74 & 35.32 & 43.45 & 46.94 \\
        \midrule
        \textbf{Avg.} & 17.16 & 21.05 & 21.24 & 21.24 & 32.07 & 40.01 & 49.52 & 52.13 & 17.97 & 19.05 & 20.08 & 20.08 
        & 22.40 & 26.70 & 30.28 & 31.15 \\
        \bottomrule
    \end{tabular}%
    }
    \caption{Percent (\%) improvement in stability loss $\mathcal{L}_S$ as function of the accuracy tolerance $\alpha$ across models and datasets. }
    \label{tab:stability_comparison}
\end{table}

\begin{table}[h]
    \centering
    \resizebox{\linewidth}{!}{%
    \begin{tabular}{lcccc|cccc|cccc|cccc}
        \toprule
        \textbf{Dataset} & \multicolumn{4}{c|}{Yearbook} & \multicolumn{4}{c|}{Healthcare} & \multicolumn{4}{c|}{Huffpost} & \multicolumn{4}{c}{\textbf{Avg.}} \\
        $\alpha$ & 0.01 & 0.02 & 0.05 & 0.1 
        & 0.01 & 0.02 & 0.05 & 0.1 
        & 0.01 & 0.02 & 0.05 & 0.1 
        & 0.01 & 0.02 & 0.05 & 0.1 \\
        \midrule
        LogReg & 0.66 & 0.66 & 0.66 & 0.66 & 0.50 & 0.70 & 0.89 & 0.89 & 0.27 & 0.27 & 0.27 & 0.27 
        & 0.48 & 0.54 & 0.61 & 0.61 \\
        CART & 0.43 & 2.15 & 1.57 & 1.57 & -0.02 & 0.31 & 0.20 & 0.20 & 1.29 & 0.78 & 2.98 & 2.98 
        & 0.56 & 1.08 & 1.58 & 1.58 \\
        XGBoost & -0.05 & -0.05 & -0.05 & -0.05 & -0.07 & 0.03 & 0.03 & 0.03 & 0.21 & 0.21 & 0.21 & 0.21 
        & 0.03 & 0.06 & 0.06 & 0.06 \\
        MLP & 0.31 & 1.18 & 1.18 & 1.18 & 1.44 & -4.51 & -0.43 & 15.01 & 0.19 & 0.19 & 0.19 & 0.19 
        & 0.65 & -1.05 & 0.31 & 4.13 \\
        \midrule
        \textbf{Avg.} & 0.34 & 0.99 & 0.84 & 0.84 & 0.46 & -0.87 & 0.17 & 4.03 & 0.49 & 0.36 & 0.91 & 0.91 
        & 0.43 & 0.16 & 0.64 & 1.93 \\
        \bottomrule
    \end{tabular}%
    }
    \caption{Percent (\%) deterioration of realized out-of-sample accuracy as a function of the selected $\alpha$ across models and datasets.}
    \label{tab:accuracy_comparison}
\end{table}

\paragraph{Robustness check: inter-sequence stability.} To test the inter-sequence stability of our methodology, we generate multiple different greedy and SVML XGBoost sequences (all with $\alpha = 0.01)$, each using one more batch of the training data. In Table \ref{table:1}, we report, for both greedy and SVML, the inter-sequence stability loss between the sequences of length $b$ and $b+1$, for each $b\in [B-1]$; e.g., $T_{1-2}$ measures the inter-sequence stability loss between $(f_1)$ and $(f_1',f_2')$, i.e., $d(f_1,f_2')$. SVML achieves significantly improved inter-sequence stability, which empirically verifies the \textit{stability by process} argument.

\begin{table}[hbt]
  \centering
  \caption{{\small Inter-sequence stability loss for varying sequence lengths}}
\resizebox{0.8\columnwidth}{!}{
  \begin{tabular}{lllllllllll}
    \toprule
    Model Name  & $T_{1-2}$ & $T_{2-3}$ & $T_{3-4}$ & $T_{4-5}$ & $T_{5-6}$ &  $T_{6-7}$ &  $T_{7-8}$ & $T_{8-9}$ & $T_{9-10}$ & $T_{10-11}$ \\
    \midrule
    \textbf{Greedy} &  \textbf{0.476} & 0.455 & 0.405 & 0.363 & 0.400 & 0.397 & 0.353 & 0.377 & \textbf{0.284} & 0.255 \\
    \textbf{SVML}  & 0.561 & \textbf{0.418} & \textbf{0.317} & \textbf{0.340} & \textbf{0.294} & \textbf{0.323} & \textbf{0.328} & \textbf{0.233} & \textbf{0.284} & \textbf{0.210} \\
    \bottomrule
  \end{tabular}
}
\label{table:1}
\end{table}

\paragraph{Robustness check: update frequency.} Next, focusing on the healthcare case study, we empirically investigate the sensitivity of SVML to different update frequencies by splitting the available 52 months of data into intervals of different lengths (1, 3, 6, and 9 months). As discussed, the currently deployed 6-month frequency will change, so our experiment also serves to understand the properties of different frequencies. We choose XGBoost as the canonical model type for the experiment since this is the model in use in production, and train 25 candidate models per batch. We find that trends and insights are consistent across the different frequencies tested. A 3-month re-training frequency exhibits more stable behavior compared to 1-month, while retaining high AUC ranging from 0.85-0.9 with improvements in distance from [0.3-0.37] to [0.275-0.21] compared to 6- or 9-month intervals. We present the detailed results in Appendix \ref{appx:additional-results}.

\paragraph{Robustness check: distance measure.} As a final robustness check, we investigate how well the analytical insights of an SVML sequence optimized for one distance measure (XGBoost with Gain importance) generalize to a different measure (SHAP) in terms of their stability across retraining iterations. Specifically, we study the empirical distribution of the absolute deviation in features' SHAP values between adjacent models for both greedy and SVML sequences. Greedy achieves a mean deviation of 0.086 with a standard deviation of 0.101 and a maximum deviation of 0.625; in comparison, SVML exhibits an improved mean deviation of 0.081 with a standard deviation of 0.094 and a maximum deviation of 0.606. These results indicate that SVML's improved stability across retraining iterations effectively translates into consistent SHAP values, affirming the method's reliability. We present the detailed results in Appendix \ref{appx:additional-results}.

\section{Conclusion} \label{s:conclude}
In summary, we have introduced a framework for retraining ML models, emphasizing structural stability across updates with new data. By leveraging a mixed-integer optimization approach, our methodology enables users to navigate the predictive power-stability Pareto frontier, and deploy stable models without significant compromises on predictive power. The practical benefits of our approach are highlighted through real-world case studies, including a production case study in a major US hospital, where consistent interpretability across retraining iterations is crucial for user trust and model adoption. Future work will look to expand the applicability of our methods to cases where distributional shifts are anticipated and can be quantified, and where cost considerations need to explicitly be modeled.

\bibliography{reference}

\begin{thebibliography}{35}
\providecommand{\natexlab}[1]{#1}
\providecommand{\url}[1]{\texttt{#1}}
\expandafter\ifx\csname urlstyle\endcsname\relax
  \providecommand{\doi}[1]{doi: #1}\else
  \providecommand{\doi}{doi: \begingroup \urlstyle{rm}\Url}\fi

\bibitem[Babic et~al.(2021)Babic, Gerke, Evgeniou, and Cohen]{babic2021beware}
Babic, B., Gerke, S., Evgeniou, T., and Cohen, I.~G.
\newblock Beware explanations from ai in health care.
\newblock \emph{Science}, 373\penalty0 (6552):\penalty0 284--286, 2021.

\bibitem[Baier et~al.(2019)Baier, J{\"o}hren, and Seebacher]{baier2019challenges}
Baier, L., J{\"o}hren, F., and Seebacher, S.
\newblock Challenges in the deployment and operation of machine learning in practice.
\newblock In \emph{ECIS}, volume~1, 2019.

\bibitem[Bargagli-Stoffi et~al.(2020)Bargagli-Stoffi, Cadei, Lee, and Dominici]{bargagli2020causal}
Bargagli-Stoffi, F.~J., Cadei, R., Lee, K., and Dominici, F.
\newblock Causal rule ensemble: Interpretable discovery and inference of heterogeneous treatment effects.
\newblock \emph{arXiv preprint arXiv:2009.09036}, 2020.

\bibitem[Basu et~al.(2018)Basu, Kumbier, Brown, and Yu]{Basu2018}
Basu, S., Kumbier, K., Brown, J.~B., and Yu, B.
\newblock Iterative random forests to discover predictive and stable high-order interactions.
\newblock \emph{Proceedings of the National Academy of Sciences}, 115\penalty0 (8):\penalty0 1943--1948, 2018.
\newblock ISSN 1091-6490.
\newblock \doi{10.1073/pnas.1711236115}.
\newblock URL \url{http://dx.doi.org/10.1073/pnas.1711236115}.

\bibitem[Bertsimas \& Digalakis~Jr(2023)Bertsimas and Digalakis~Jr]{stable_tree}
Bertsimas, D. and Digalakis~Jr, V.
\newblock Improving stability in decision tree models.
\newblock \emph{arXiv preprint arXiv:2305.17299}, 2023.
\newblock \doi{10.48550/ARXIV.2305.17299}.

\bibitem[Bertsimas \& Orfanoudaki(2021)Bertsimas and Orfanoudaki]{bertsimas2021algorithmic}
Bertsimas, D. and Orfanoudaki, A.
\newblock Algorithmic insurance.
\newblock \emph{arXiv preprint arXiv:2106.00839}, 2021.

\bibitem[Bertsimas et~al.(2024)Bertsimas, Digalakis~Jr, Li, and Lami]{bertsimas2024slowly}
Bertsimas, D., Digalakis~Jr, V., Li, M.~L., and Lami, O.~S.
\newblock Slowly varying regression under sparsity.
\newblock \emph{Operations Research}, 2024.

\bibitem[Bifet \& Gavalda(2007)Bifet and Gavalda]{bifet2007learning}
Bifet, A. and Gavalda, R.
\newblock Learning from time-changing data with adaptive windowing.
\newblock In \emph{Proceedings of the 2007 SIAM international conference on data mining}, pp.\  443--448. SIAM, 2007.

\bibitem[Bisong(2019)]{Bisong2019}
Bisong, E.
\newblock \emph{Batch vs. Online Learning}, pp.\  199--201.
\newblock Apress, Berkeley, CA, 2019.
\newblock ISBN 978-1-4842-4470-8.
\newblock \doi{10.1007/978-1-4842-4470-8_15}.
\newblock URL \url{https://doi.org/10.1007/978-1-4842-4470-8_15}.

\bibitem[Bousquet \& Elisseeff(2002)Bousquet and Elisseeff]{bousquet2002stability}
Bousquet, O. and Elisseeff, A.
\newblock Stability and generalization.
\newblock \emph{The Journal of Machine Learning Research}, 2:\penalty0 499--526, 2002.

\bibitem[Breiman(1996)]{breiman1996heuristics}
Breiman, L.
\newblock Heuristics of instability and stabilization in model selection.
\newblock \emph{The Annals of Statistics}, 24\penalty0 (6):\penalty0 2350--2383, 1996.

\bibitem[Breiman et~al.(1984)Breiman, Friedman, Olshen, and Stone]{breiman1984classification}
Breiman, L., Friedman, J., Olshen, R., and Stone, C.
\newblock \emph{Classification and regression trees}.
\newblock Monterey, CA: Wadsworth and Brooks, 1984.

\bibitem[Cortes et~al.(2020)Cortes, Mohri, Gonzalvo, and Storcheus]{cortes2020agnostic}
Cortes, C., Mohri, M., Gonzalvo, J., and Storcheus, D.
\newblock Agnostic learning with multiple objectives.
\newblock \emph{Advances in Neural Information Processing Systems}, 33:\penalty0 20485--20495, 2020.

\bibitem[Dietvorst et~al.(2018)Dietvorst, Simmons, and Massey]{dietvorst2018overcoming}
Dietvorst, B.~J., Simmons, J.~P., and Massey, C.
\newblock Overcoming algorithm aversion: People will use imperfect algorithms if they can (even slightly) modify them.
\newblock \emph{Management Science}, 64\penalty0 (3):\penalty0 1155--1170, 2018.

\bibitem[Dwyer \& Holte(2007)Dwyer and Holte]{Dwyer_2007}
Dwyer, K. and Holte, R.
\newblock Decision tree instability and active learning.
\newblock In \emph{Machine Learning: ECML 2007: 18th European Conference on Machine Learning, Warsaw, Poland, September 17-21, 2007. Proceedings 18}, pp.\  128--139. Springer, 2007.

\bibitem[Gama et~al.(2014)Gama, {\v{Z}}liobait{\.e}, Bifet, Pechenizkiy, and Bouchachia]{gama2014survey}
Gama, J., {\v{Z}}liobait{\.e}, I., Bifet, A., Pechenizkiy, M., and Bouchachia, A.
\newblock A survey on concept drift adaptation.
\newblock \emph{ACM computing surveys (CSUR)}, 46\penalty0 (4):\penalty0 1--37, 2014.

\bibitem[Glikson \& Woolley(2020)Glikson and Woolley]{glikson2020human}
Glikson, E. and Woolley, A.~W.
\newblock Human trust in artificial intelligence: Review of empirical research.
\newblock \emph{Academy of Management Annals}, 14\penalty0 (2):\penalty0 627--660, 2020.

\bibitem[Hoi et~al.(2021)Hoi, Sahoo, Lu, and Zhao]{hoi2021online}
Hoi, S.~C., Sahoo, D., Lu, J., and Zhao, P.
\newblock Online learning: A comprehensive survey.
\newblock \emph{Neurocomputing}, 459:\penalty0 249--289, 2021.

\bibitem[Kabra \& Patel(2024)Kabra and Patel]{kabra2024limitations}
Kabra, A. and Patel, K.~K.
\newblock The limitations of model retraining in the face of performativity.
\newblock \emph{arXiv preprint arXiv:2408.08499}, 2024.

\bibitem[Longoni et~al.(2019)Longoni, Bonezzi, and Morewedge]{longoni2019resistance}
Longoni, C., Bonezzi, A., and Morewedge, C.~K.
\newblock Resistance to medical artificial intelligence.
\newblock \emph{Journal of Consumer Research}, 46\penalty0 (4):\penalty0 629--650, 2019.

\bibitem[Lundberg \& Lee(2017)Lundberg and Lee]{shap}
Lundberg, S.~M. and Lee, S.-I.
\newblock A unified approach to interpreting model predictions.
\newblock In \emph{Proceedings of the 31st International Conference on Neural Information Processing Systems}, NIPS'17, pp.\  4768--4777, Red Hook, NY, USA, 2017. Curran Associates Inc.
\newblock ISBN 9781510860964.

\bibitem[Mahadevan \& Mathioudakis(2023)Mahadevan and Mathioudakis]{mahadevan2023cost}
Mahadevan, A. and Mathioudakis, M.
\newblock Cost-effective retraining of machine learning models.
\newblock \emph{arXiv preprint arXiv:2310.04216}, 2023.

\bibitem[Miettinen(1999)]{miettinen1999nonlinear}
Miettinen, K.
\newblock \emph{Nonlinear multiobjective optimization}, volume~12.
\newblock Springer Science \& Business Media, 1999.

\bibitem[Orfanoudaki et~al.(2022)Orfanoudaki, Saghafian, Song, Chakkera, and Cook]{Orfanoudaki2022}
Orfanoudaki, A., Saghafian, S., Song, K., Chakkera, H.~A., and Cook, C.
\newblock Algorithm, human, or the centaur: How to enhance clinical care?
\newblock \emph{SSRN Electronic Journal}, 2022.
\newblock ISSN 1556-5068.
\newblock \doi{10.2139/ssrn.4302002}.
\newblock URL \url{http://dx.doi.org/10.2139/ssrn.4302002}.

\bibitem[Pedregosa et~al.(2011)Pedregosa, Varoquaux, Gramfort, Michel, Thirion, Grisel, Blondel, Prettenhofer, Weiss, Dubourg, et~al.]{pedregosa2011scikit}
Pedregosa, F., Varoquaux, G., Gramfort, A., Michel, V., Thirion, B., Grisel, O., Blondel, M., Prettenhofer, P., Weiss, R., Dubourg, V., et~al.
\newblock Scikit-learn: Machine learning in python.
\newblock \emph{Journal of Machine Learning Research}, 12:\penalty0 2825--2830, 2011.

\bibitem[Pesaranghader \& Viktor(2016)Pesaranghader and Viktor]{pesaranghader2016fast}
Pesaranghader, A. and Viktor, H.~L.
\newblock Fast hoeffding drift detection method for evolving data streams.
\newblock In \emph{Machine Learning and Knowledge Discovery in Databases: European Conference, ECML PKDD 2016, Riva del Garda, Italy, September 19-23, 2016, Proceedings, Part II 16}, pp.\  96--111. Springer, 2016.

\bibitem[Raghu et~al.(2017)Raghu, Gilmer, Yosinski, and Sohl-Dickstein]{raghu2017svcca}
Raghu, M., Gilmer, J., Yosinski, J., and Sohl-Dickstein, J.
\newblock Svcca: Singular vector canonical correlation analysis for deep learning dynamics and interpretability.
\newblock \emph{Advances in neural information processing systems}, 30, 2017.

\bibitem[Schwinn et~al.(2022)Schwinn, Bungert, Nguyen, Raab, Pulsmeyer, Precup, Eskofier, and Zanca]{schwinn2022improving}
Schwinn, L., Bungert, L., Nguyen, A., Raab, R., Pulsmeyer, F., Precup, D., Eskofier, B., and Zanca, D.
\newblock Improving robustness against real-world and worst-case distribution shifts through decision region quantification.
\newblock In \emph{International Conference on Machine Learning}, pp.\  19434--19449. PMLR, 2022.

\bibitem[S{\'u}ken{\'\i}k \& Lampert(2024)S{\'u}ken{\'\i}k and Lampert]{sukenik2022generalization}
S{\'u}ken{\'\i}k, P. and Lampert, C.
\newblock Generalization in multi-objective machine learning.
\newblock \emph{Neural Computing and Applications}, pp.\  1--15, 2024.

\bibitem[Turney(1995)]{turney1995bias}
Turney, P.
\newblock Bias and the quantification of stability.
\newblock \emph{Machine Learning}, 20:\penalty0 23--33, 1995.

\bibitem[Wang et~al.(2024)Wang, Wu, and Nettleton]{wang2024stability}
Wang, Y., Wu, H., and Nettleton, D.
\newblock Stability of random forests and coverage of random-forest prediction intervals.
\newblock \emph{Advances in Neural Information Processing Systems}, 36, 2024.

\bibitem[Wu et~al.(2020)Wu, Dobriban, and Davidson]{wu2020deltagrad}
Wu, Y., Dobriban, E., and Davidson, S.
\newblock Deltagrad: Rapid retraining of machine learning models.
\newblock In \emph{International Conference on Machine Learning}, pp.\  10355--10366. PMLR, 2020.

\bibitem[Yao et~al.(2022)Yao, Choi, Cao, Lee, Koh, and Finn]{yao2022wild}
Yao, H., Choi, C., Cao, B., Lee, Y., Koh, P. W.~W., and Finn, C.
\newblock Wild-time: A benchmark of in-the-wild distribution shift over time.
\newblock \emph{Advances in Neural Information Processing Systems}, 35:\penalty0 10309--10324, 2022.

\bibitem[Zhang et~al.(2020)Zhang, Feng, Wang, He, Wang, Li, and Zhang]{zhang2020retrain}
Zhang, Y., Feng, F., Wang, C., He, X., Wang, M., Li, Y., and Zhang, Y.
\newblock How to retrain recommender system? a sequential meta-learning method.
\newblock In \emph{Proceedings of the 43rd International ACM SIGIR Conference on Research and Development in Information Retrieval}, pp.\  1479--1488, 2020.

\bibitem[{\v{Z}}liobait{\.e} et~al.(2015){\v{Z}}liobait{\.e}, Budka, and Stahl]{vzliobaite2015towards}
{\v{Z}}liobait{\.e}, I., Budka, M., and Stahl, F.
\newblock Towards cost-sensitive adaptation: When is it worth updating your predictive model?
\newblock \emph{Neurocomputing}, 150:\penalty0 240--249, 2015.

\end{thebibliography}
\bibliographystyle{icml2025}

\newpage
\appendix
\onecolumn
\section{Technical Proofs} \label{appx:technical-proofs}
\begin{proof}{(Proof of Theorem \ref{theo:wpo})}  
    Let $\boldsymbol{f^*}$ be a solution to Problem \eqref{eqn:formulation-constrained}. Assume $\boldsymbol{f^*}$ is not WPO. Then, $\exists \boldsymbol{f'}$ such that $\hat{\mathcal{L}}_P(f'_b, D_b) < \hat{\mathcal{L}}_P(f^*_b, D_b) \leq \epsilon_b \ \forall b$ (so $\boldsymbol{f'}$ is feasible) and $\hat{\mathcal{L}}_S(\boldsymbol{f'}) < \hat{\mathcal{L}}_S(\boldsymbol{f^*})$. This contradicts the optimality of $\boldsymbol{f^*}$ among feasible $\boldsymbol{f}$; thus, $\nexists$ such $\boldsymbol{f'}$ so $\boldsymbol{f^*}$ is WPO.
\end{proof}

\begin{proof}{(Proof of Theorem \ref{theo:po})}
    Let $\boldsymbol{f^*}$ be a unique solution to Problem \eqref{eqn:formulation-constrained}. Then, $\hat{\mathcal{L}}_S(\boldsymbol{f^*}) < \hat{\mathcal{L}}_S(\boldsymbol{f}) \ \forall \boldsymbol{f}$ when $\hat{\mathcal{L}}_P(f_b, D_b) \leq \epsilon_b \ \forall b$. Assume $\boldsymbol{f^*}$ is not PO. Then, $\exists \boldsymbol{f'}$ such that $\hat{\mathcal{L}}_P(f'_b, D_b) \leq \hat{\mathcal{L}}_P(f^*_b, D_b)\ \forall b$, $\hat{\mathcal{L}}_S(\boldsymbol{f'}) \leq \hat{\mathcal{L}}_S(\boldsymbol{f^*})$, and the inequality is strict for at least one $b' \in [B]$ or for the stability loss. If the inequality is strict for the stability loss, then we have $\hat{\mathcal{L}}_S(\boldsymbol{f'}) < \hat{\mathcal{L}}_S(\boldsymbol{f^*})$ (so $\boldsymbol{f'}$ achieves a better objective value in Problem \eqref{eqn:formulation-constrained}) and $\hat{\mathcal{L}}_P(f'_b, D_b) \leq \hat{\mathcal{L}}_P(f^*_b, D_b) \leq \epsilon_b\ \forall b$ (so $\boldsymbol{f'}$ is feasible for Problem \eqref{eqn:formulation-constrained}). Thus, we have a contradiction with the fact that $\boldsymbol{f^*}$ solves Problem \eqref{eqn:formulation-constrained}. If the inequality is strict for $b' \in [B]$, then $\hat{\mathcal{L}}_P(f'_{b'}, D_{b'}) < \hat{\mathcal{L}}_P(f^*_{b'}, D_{b'}) \leq \epsilon_{b'}$, $\hat{\mathcal{L}}_P(f'_b, D_b) \leq \hat{\mathcal{L}}_P(f^*_b, D_b) \leq \epsilon_b\ \forall b \not= b'$, and $\hat{\mathcal{L}}_S(\boldsymbol{f'}) \leq \hat{\mathcal{L}}_S(\boldsymbol{f^*})$.
    This contradicts the uniqueness of $\boldsymbol{f^*}$. Thus, $\nexists$ such $\boldsymbol{f'}$ and $\boldsymbol{f^*}$ is PO.
\end{proof}

\begin{proof}{(Proof of Theorem \ref{theo:pov})}
    Let $\boldsymbol{f^*}$ be a solution to Problem \eqref{eqn:formulation-constrained} with $\epsilon_b = \epsilon^*_b\ \forall b$ and the complementary Problem \eqref{eqn:formulation-complementary} $\forall b' \in [B]$. Then, by optimality of $\boldsymbol{f^*}$ for Problem \eqref{eqn:formulation-constrained}, $\nexists \boldsymbol{f'}$ such that $\hat{\mathcal{L}}_S(\boldsymbol{f'}) < \hat{\mathcal{L}}_S(\boldsymbol{f}^*)$ and $\hat{\mathcal{L}}_P(f'_b, D_b) \leq \epsilon^*_b\ \forall b$. Further, by optimality of $\boldsymbol{f^*}$ for Problem \eqref{eqn:formulation-complementary} for all $b' \in [B]$, $\nexists \boldsymbol{f'}$ such that $\hat{\mathcal{L}}_P(f'_{b'}, D_{b'}) < \epsilon^*_{b'}$, $\hat{\mathcal{L}}_S(\boldsymbol{f'}) \leq \hat{\mathcal{L}}_S(\boldsymbol{f}^*)$, and $\hat{\mathcal{L}}_P(f'_b, D_b) \leq \epsilon^*_b\ \forall b\not=b'$. Thus, $\boldsymbol{f^*}$ is PO since, in both cases, the inequality is strict for at least one objective.

    For the opposite direction, let $\boldsymbol{f^*}$ be PO. First, assume $\boldsymbol{f^*}$ does not solve Problem \eqref{eqn:formulation-constrained} with $\epsilon_b = \epsilon^*_b\ \forall b$. Then, $\exists \boldsymbol{f'}$ that is feasible, i.e., $\hat{\mathcal{L}}_P(f'_b, D_b) \leq \epsilon^*_b\ \forall b$, such that $\hat{\mathcal{L}}_S(\boldsymbol{f'}) < \hat{\mathcal{L}}_S(\boldsymbol{f}^*)$. This contradicts the PO of $\boldsymbol{f^*}$. Now, assume $\boldsymbol{f^*}$ does not solve Problem \eqref{eqn:formulation-complementary} for some $b'$. Then, $\exists \boldsymbol{f'}$ that is feasible, i.e., $\hat{\mathcal{L}}_P(f'_b, D_b) \leq \epsilon^*_b\ \forall b \not= b'$ and $\hat{\mathcal{L}}_S(\boldsymbol{f'}) \leq \hat{\mathcal{L}}_S(\boldsymbol{f}^*)$, such that $\hat{\mathcal{L}}_P(f'_{b'}, D_b) < \epsilon^*_{b'}$. This also contradicts the PO of $\boldsymbol{f^*}$. Thus, $\boldsymbol{f^*}$ solves the series of Problems stated in Theorem \ref{theo:pov}.
\end{proof}

\begin{proof}{(Proof of Theorem \ref{theo:shortest-path-reduction})}
    Construct a directed graph \( G\) as follows: Each node \( v_{b, j}\) corresponds to model \( f_{b, j} \in \tilde{\mathcal{F}}_b\) ($j$-th model for batch \( b \)).
    Include additional nodes \( v_{\text{source}} \) and \( v_{\text{sink}} \). 
    Build directed edges from \( v_{\text{source}} \) to \(v_{1,j}, \forall j,\) and from \(v_{B,j}\) to \( v_{\text{sink}}, \forall j\), with zero weights. For $b \in [B-1]$, build directed edges from \(v_{b,j}\) to \(v_{b+1,k}, \forall (k,j)\); assign a weight of \( d(f_{b, j}, f_{b+1, k}) \) to each edge \( (v_{b, j}, v_{b+1, k}) \). 
    To enforce the accuracy tolerance constraint, for each batch \( b \), include only those nodes \( v_{b, j} \) corresponding to models \( f_{b, j} \) that satisfy \( \mathcal{L}_P(f_{b, j}, D_b) \leq (1 + \alpha) \cdot \min_{f \in \mathcal{F}_b} \mathcal{L}_P(f, D_b) \).
    Finding a sequence of models that minimizes the total stability metric while satisfying the accuracy tolerance constraint is equivalent to finding the shortest path from \( v_{\text{source}} \) to \( v_{\text{sink}} \) in graph \( G \).
    Since the shortest path problem in a directed graph with non-negative weights can be solved in polynomial time (e.g., using Dijkstra's algorithm), Problem \eqref{eqn:formulation-restricted} can also be solved in polynomial time.
\end{proof}

\begin{proof}{(Proof of Theorem \ref{theo:intra-inter})}
The result follows by repeated application of the triangle inequality:
\begin{equation*}
    \begin{split}
        \mathcal{L}_S(\boldsymbol{f}, \boldsymbol{f’}) 
        & := d(f_B, f’_{B+1}) 
          \leq d(f_{B-1}, f_B) + d(f_{B-1}, f’_{B+1}) 
          \leq \dots \leq \sum_{b=1}^{B-1} d(f_b, f_{b+1}) + d(f_{1}, f’_{B+1}) \\
        & \leq \dots \leq \mathcal{L}_S(\boldsymbol{f}) + \sum_{b=1}^{B} d(f’_b, f’_{b+1})  + d(f_1, f’_1) 
          \leq \mathcal{L}_S(\boldsymbol{f}) + \mathcal{L}_S(\boldsymbol{f’}) + d(f_0, f_0) = \epsilon + \epsilon’.
    \end{split}
\end{equation*}
\end{proof}

\section{Numerical Experiments on the Pareto Excess Risk Bound for Linear Regression} \label{appx:excess-risk-bound}
We numerically verify the finite-sample high-probability excess risk bounds for linear regression for both independent training and multi-objective stable training across increasing batch sizes. First, in Figure~\ref{fig:excess_risk_bounds}, we plot the theoretical bounds derived in Section~\ref{sec:problem-statement}. Specifically, for each batch size \( N_b \), we compute:
\begin{itemize}
    \item The independent training bound, given by:
    \begin{equation} \label{eq:independent-bound}
        \mathcal{L}_P(\hat{f}_b, D_b) - \mathcal{L}_P(f^*_b, D_b) \leq \frac{\sigma^2 d}{N_b} + 2 \sigma \sqrt{\frac{4 d \log(1/\delta)}{N_b}}.
    \end{equation}
    \item The Pareto-optimal bound, which accounts for stability across all \( B \) batches:
    \begin{equation} \label{eq:pareto-bound}
        \mathcal{L}_P(\hat{f^*_b}, D_b) - \mathcal{L}_P(f^*_b, D_b) \leq 2 \frac{\sigma^2 d}{N_b} + \sigma \sqrt{\frac{4 d \log(B/\delta)}{N_b}}, \quad \forall b \in [B].
    \end{equation}
\end{itemize}
The plots show these bounds as functions of the batch size \( N_b \) for different total numbers of batches: on the left, for \( B = 5 \); on the right, for \( B = 10 \).

\begin{figure}[h]
    \centering
    \includegraphics[width=0.8\textwidth]{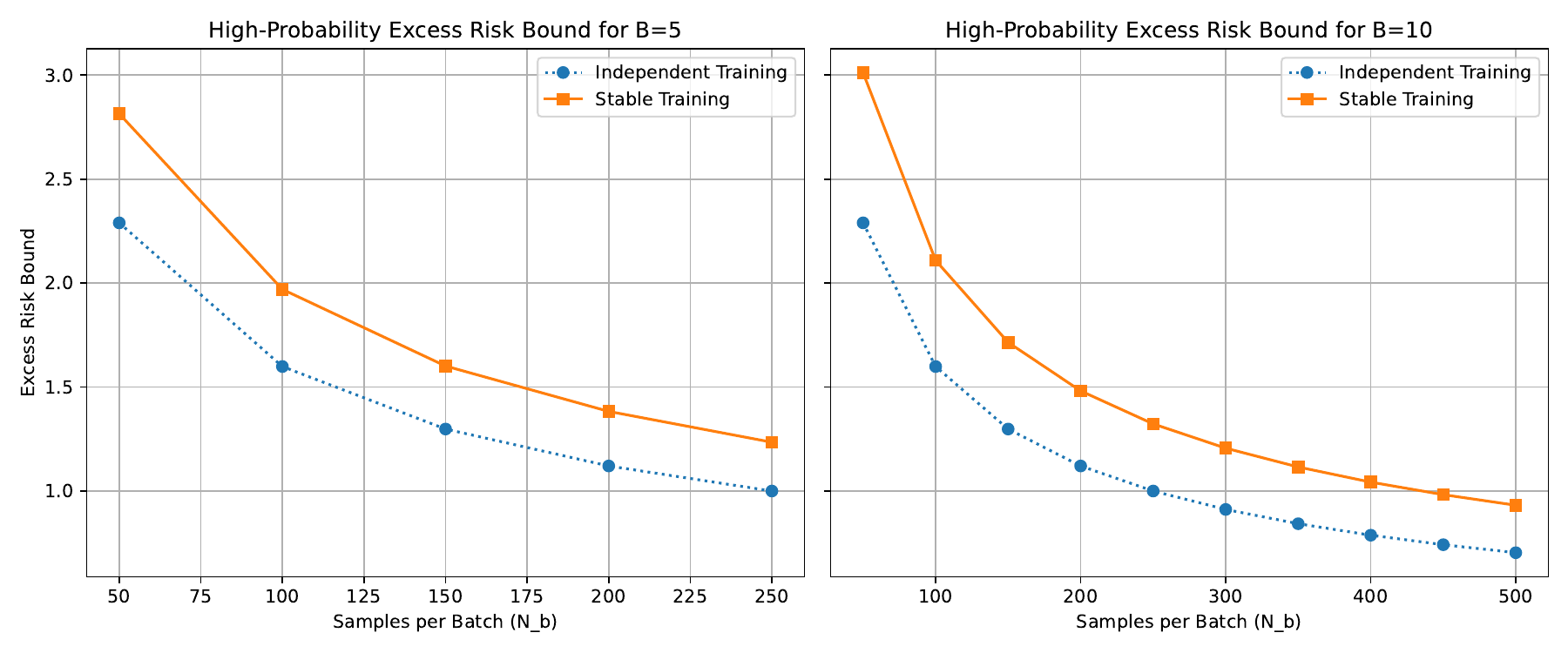}
    \caption{High-probability excess risk bound for linear regression across batches. The dotted line represents the independent bound \eqref{eq:independent-bound}, while the solid line represents the Pareto-optimal bound \eqref{eq:pareto-bound}. Left: \( B = 5 \); Right: \( B = 10 \).}
    \label{fig:excess_risk_bounds}
\end{figure}

\paragraph{Experimental Setup.} 
For each batch \( b \in [B] \), we generate a dataset \( D_b = \{(\boldsymbol{x}_i, y_i)\}_{i=1}^{N_b} \) where \( \boldsymbol{x}_i \sim \mathcal{N}(0, I_p) \) and the ground-truth model follows \( f^*_b(\boldsymbol{x}) = \boldsymbol{x}^\top \boldsymbol{\beta}_b \), with \( \boldsymbol{\beta}_b \sim \mathcal{N}(0, I_p) \). The response is \( y_i = f^*_b(\boldsymbol{x}_i) + \epsilon_i \), with \( \epsilon_i \sim \mathcal{N}(0, \sigma^2) \). We set \( p = 5 \), \( \sigma^2 = 1 \), and vary \( N_b \) from \( 100 \) to \( 2000 \). The number of batches is \( B \in \{5, 10\} \), and confidence level \( \delta = 0.1 \). Each experiment is repeated for \( 100 \) Monte Carlo trials.

We train an OLS model per batch and compute its empirical loss:
\begin{equation}
    \hat{\mathcal{L}}_P(f_b, D_b) = \frac{1}{N_b} \sum_{i=1}^{N_b} (y_i - f_b(\boldsymbol{x}_i))^2.
\end{equation}
We plug this into \eqref{eq:independent-bound} to evaluate the empirical probability that the bound holds for each model individually; then, we plug this into \eqref{eq:pareto-bound} to evaluate the empirical probability that the bound holds for the entire sequence.

\paragraph{Results.} 
Figure~\ref{fig:excess_risk_bounds_empirical} presents the empirical verification of the bounds. For each \( N_b \), we compute the fraction of trials where the empirical loss remains within the theoretical bound, both per batch (independent) and across all batches (Pareto). As expected:
\begin{itemize}
    \item The independent bound holds with increasing frequency as \( N_b \) grows.
    \item The Pareto bound is more conservative, as it requires all batches to satisfy their bounds simultaneously.
    \item Increasing \( B \) makes the Pareto bound harder to satisfy, but the empirical Pareto frontier converges to the true Pareto frontier as \( N_b \) increases.
\end{itemize}
In all, the results confirm that while stability-regularized training introduces a small increase in excess risk, it ensures smoother transitions while maintaining Pareto optimality across models.
\begin{figure}[h]
    \centering
    \includegraphics[width=\textwidth]{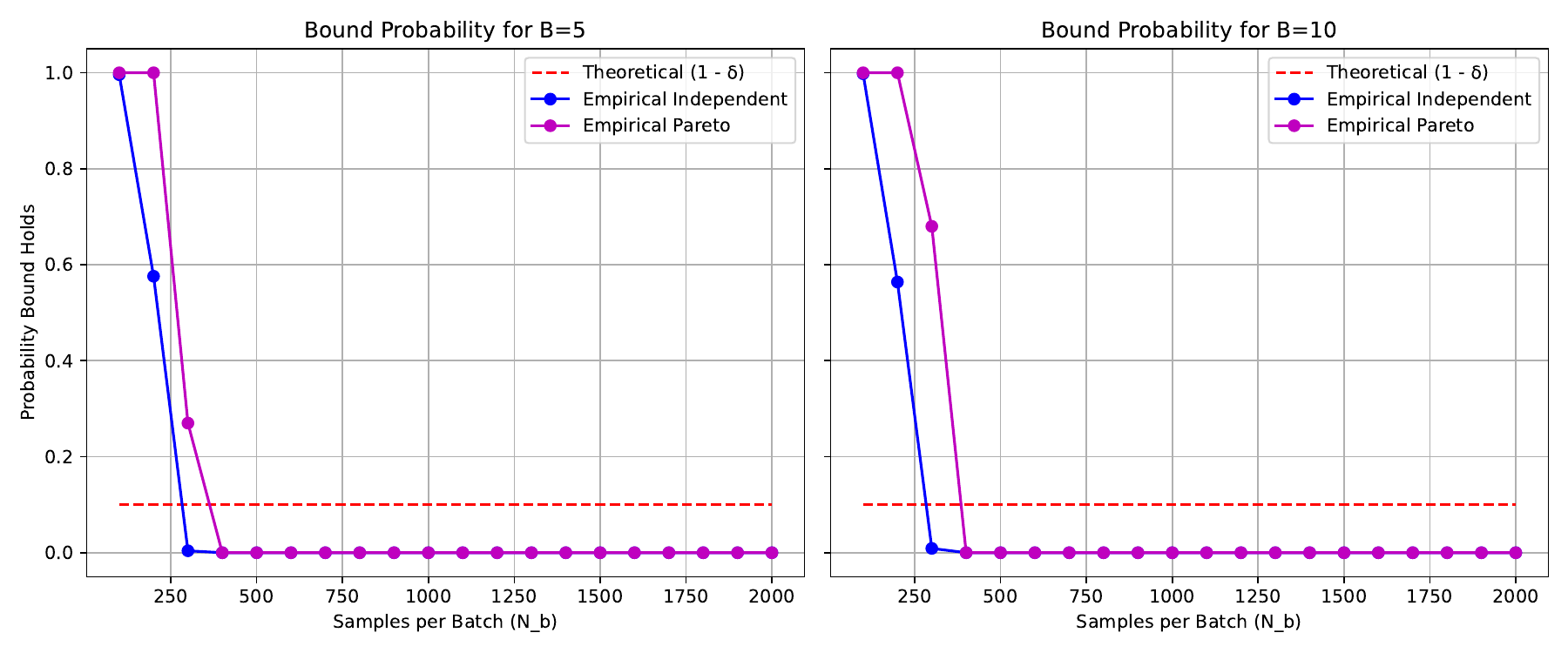}
    \caption{Empirical validation of excess risk bounds.}
    \label{fig:excess_risk_bounds_empirical}
\end{figure}

\section{Numerical Experiments on Measuring the Stability Loss} \label{appx:distance-experiment}
To evaluate how different stability loss metrics behave and how structural stability affects analytical insights and predictions, we conduct a controlled experiment across three model families: logistic regression, decision trees, and neural networks. We train three models on variations of the same dataset and analyze their structural distance, analytical insight-based distance, and predictive distance (output differences).

\textbf{Experimental Setup.}
We generate a synthetic binary classification dataset with 20 features, of which 10 are informative. The dataset remains the same across models, but feature availability is manipulated as follows:
\begin{itemize}
    \item \textbf{Model 1 ($f_1$) and Model 2 ($f_2$)} are trained on different bootstrap samples of the same half of the features.
    \item \textbf{Model 3 ($f_3$)} is trained on the complementary half of the features.
\end{itemize}
By training $f_1$ and $f_2$ on overlapping feature sets while $f_3$ is trained on a disjoint set, we separate the effects of feature selection from other sources of variation and analyze how changes in structure affect analytical insights and predictions.

For each model, we compute:
\begin{itemize}
    \item \textbf{Structural distance} ($d(f_1, f_2)$ and $d(f_1, f_3)$): Differences in model parameters.
    \item \textbf{Analytical insight-based distance} ($d_\phi(f_1, f_2)$ and $d_\phi(f_1, f_3)$): Differences in feature importances.
    \item \textbf{Predictive distance} ($d_{\text{pred}}(f_1, f_2)$ and $d_{\text{pred}}(f_1, f_3)$): Differences in model predictions.
\end{itemize}
Each experiment is repeated 10 times, and results are averaged.

\textbf{Results.}
Table~\ref{tab:model_stability_results} summarizes the results across all model families, whereas Figures~\ref{fig:distance_logistic}, \ref{fig:distance_tree}, and \ref{fig:distance_nn} illustrate the decision boundaries and feature importances for one random instance of the experiment for each model family.
\begin{table}[h]
    \centering
    \caption{Stability Distance Metrics Across Model Families (Averaged over 10 Runs)}
    \label{tab:model_stability_results}
    \resizebox{\textwidth}{!}{%
    \begin{tabular}{lccccccccc}
        \toprule
        \textbf{Model Type} & \textbf{$d(f_1, f_2)$} & \textbf{$d(f_1, f_3)$} & \textbf{$d_\phi(f_1, f_2)$} & \textbf{$d_\phi(f_1, f_3)$} & \textbf{$d_{\text{pred}}(f_1, f_2)$} & \textbf{$d_{\text{pred}}(f_1, f_3)$} & {AUC ($f_1$)} & {AUC ($f_2$)} & {AUC ($f_3$)} \\
        \midrule
        Logistic Regression & 0.021 & 0.118 & 0.378 & 2.356 & 0.081 & 0.395 & 0.790 & 0.795 & 0.780 \\
        Decision Trees & 0.083 & 0.192 & 0.188 & 0.719 & 0.239 & 0.435 & 0.737 & 0.755 & 0.699 \\
        Neural Networks & 1.047 & 1.387 & 0.062 & 0.300 & 0.153 & 0.375 & 0.860 & 0.854 & 0.822 \\
        \bottomrule
    \end{tabular}%
    }
\end{table}

\begin{figure}[htbp]
    \centering
    \includegraphics[width=0.7\textwidth]{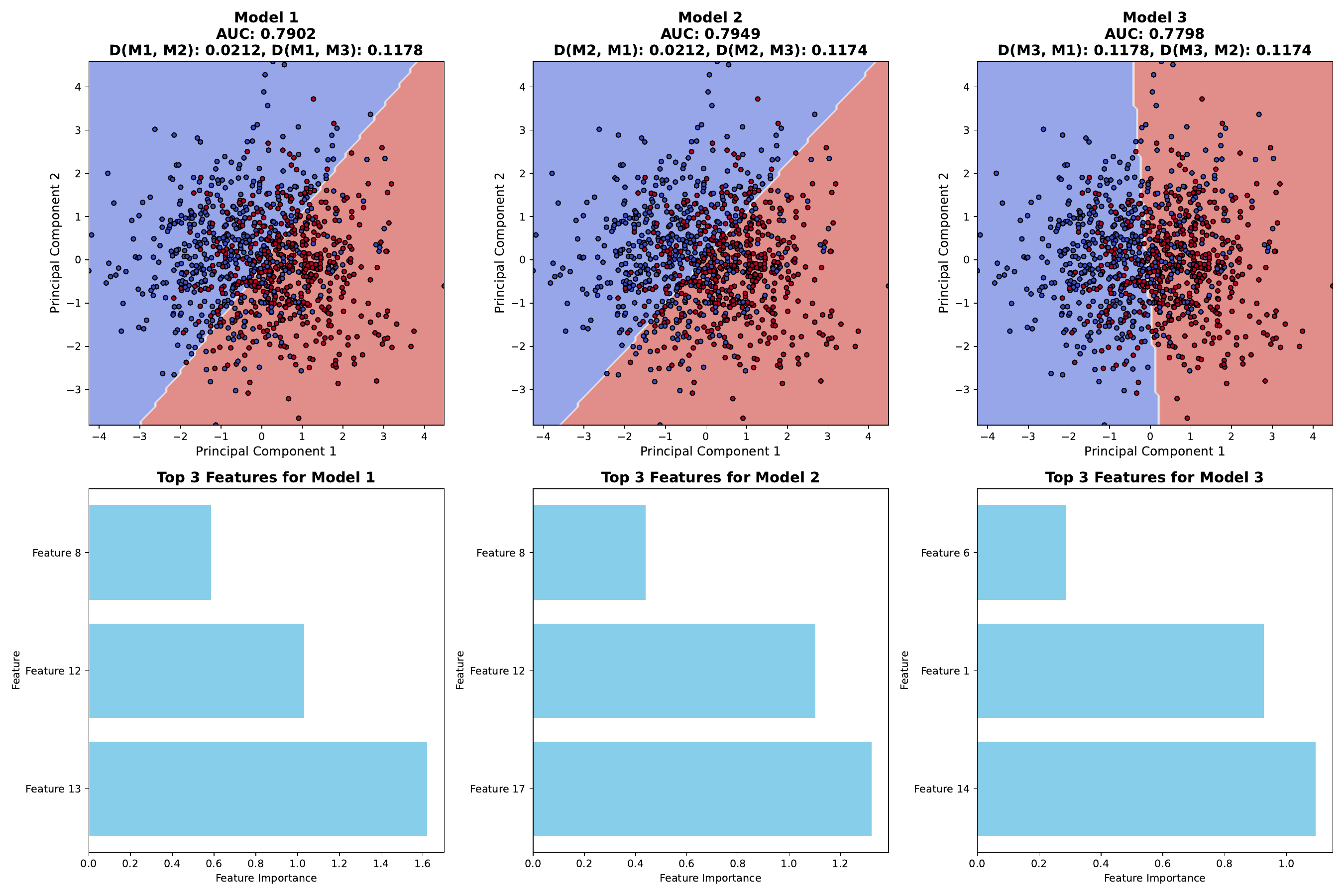}
    \caption{Decision boundaries (top) and top feature importances (bottom) for logistic regression models.}
    \label{fig:distance_logistic}
\end{figure}

\begin{figure}[htbp]
    \centering
    \includegraphics[width=0.7\textwidth]{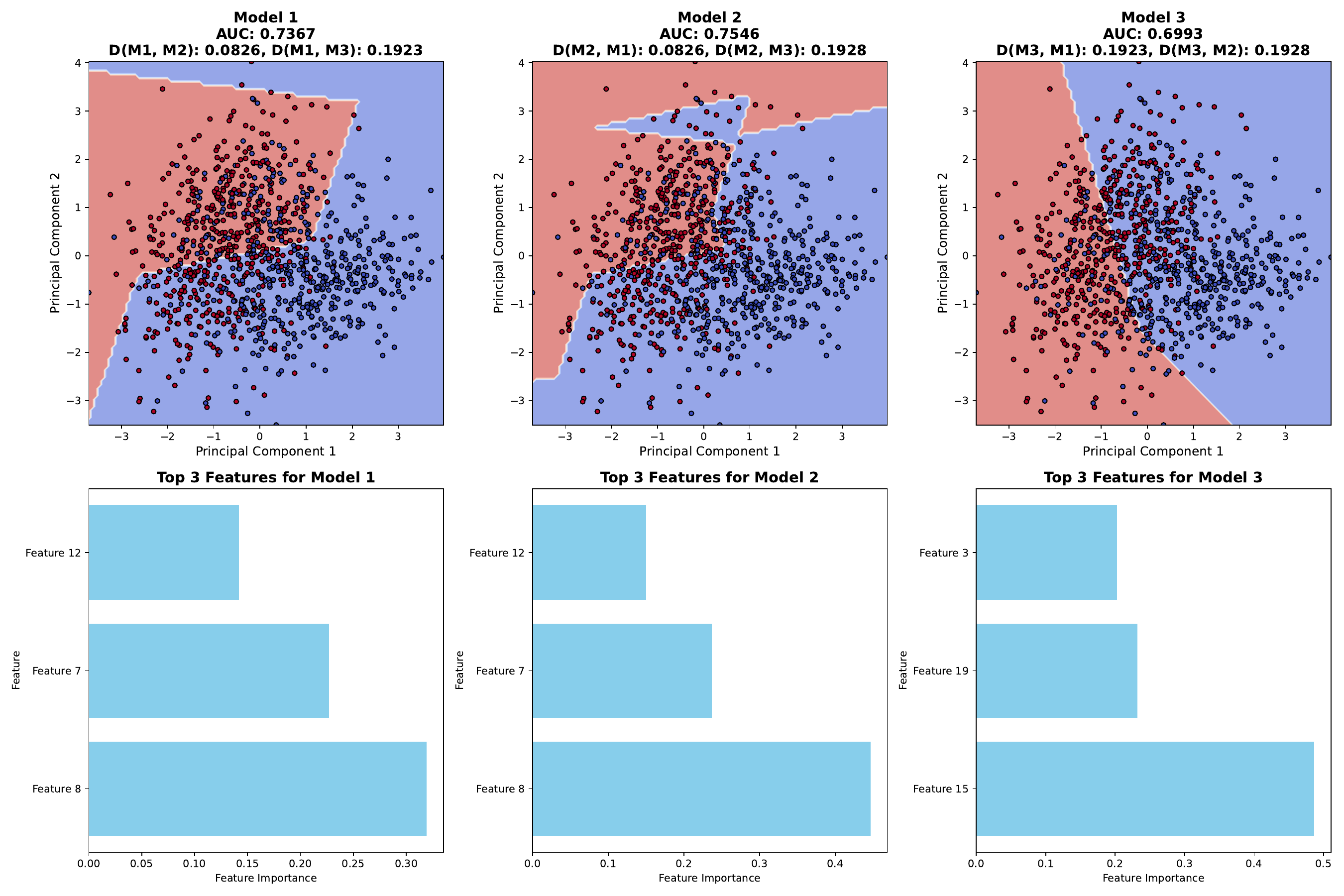}
    \caption{Decision boundaries (top) and top feature importances (bottom) for decision tree models.}
    \label{fig:distance_tree}
\end{figure}

\begin{figure}[htbp]
    \centering
    \includegraphics[width=0.7\textwidth]{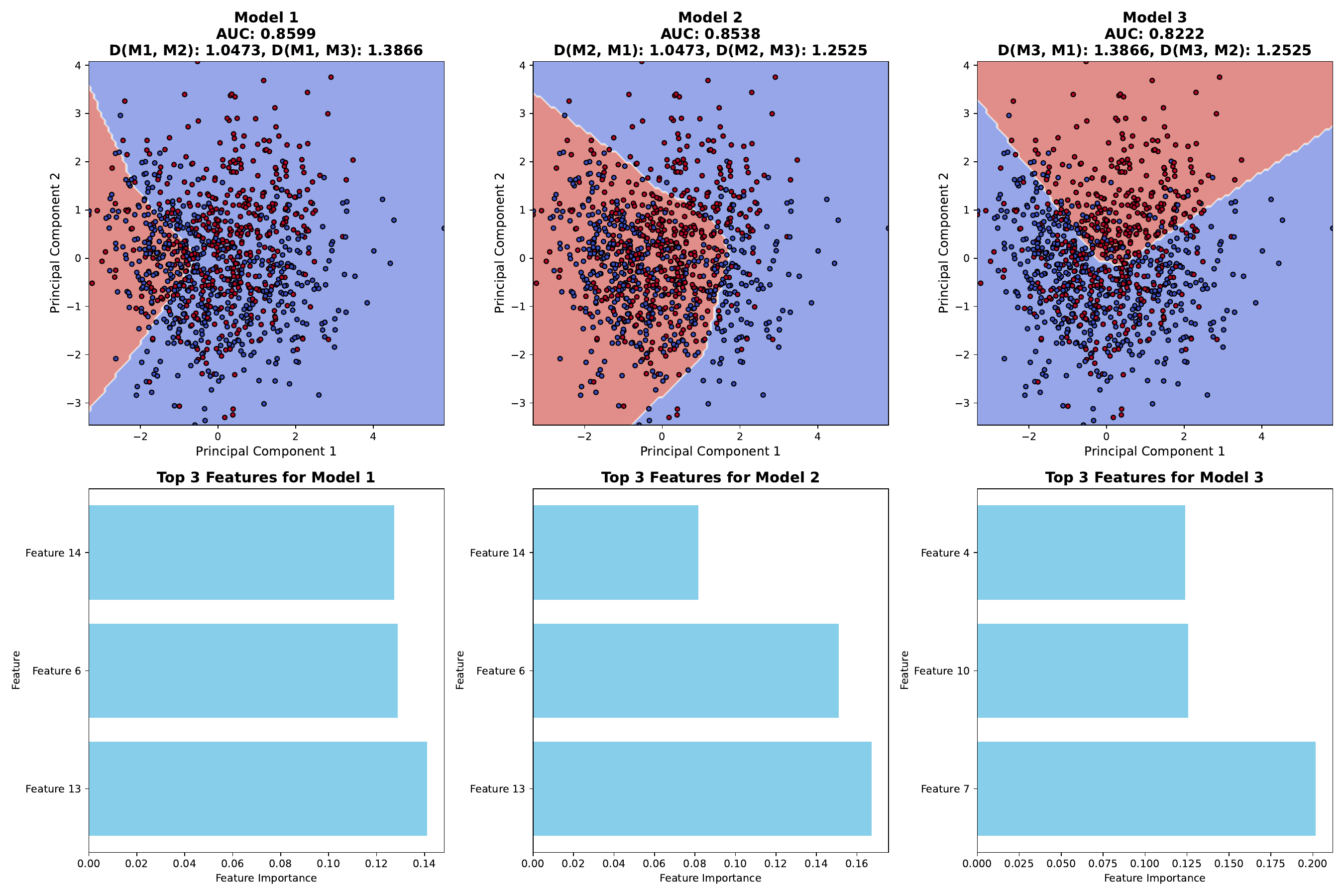}
    \caption{Decision boundaries (top) and top feature importances (bottom) for neural network models.}
    \label{fig:distance_nn}
\end{figure}

We summarize the main findings from our experiment:
\begin{itemize}
    \item \textbf{Structural distances correlate with analytical insight-based distances.} Logistic regression and decision trees exhibit strong alignment between structural stability and feature importance stability. However, neural networks show lower analytical sensitivity: despite large changes in learned parameters, the feature importance distances remain lower compared to other model families.
    \item \textbf{Structural distances correlate with predictive distances.} Across all models, larger parameter shifts correspond to greater predictive changes, particularly for decision trees and neural networks. 
    \item \textbf{Feature shifts translate into prediction instability.} Across all models, predictive differences increase with both structural and analytical instability, confirming that feature selection plays a central role in maintaining prediction consistency.
\end{itemize}

\newpage
\section{Additional Results for Section \ref{s:experiments}} \label{appx:additional-results}

\paragraph{Experiment hyperparameters.}
We outline the exact parameter search space of each of the four ML models we have run. 
\begin{itemize}
    \item For XGBoost, we do not use label encoder, AUC as the evaluation metric, 100 estimators, learning rate of 0.1, max depth of 6, objective of binary logistic or multi softmax, and scale pos weight of 100. 
    \item For logistic regression, we use balanced class weight.
    \item For CART, we use a maximum depth of 6, minimum number of samples per leaf of 50, and balanced class weight.
    \item For MLP, we used hidden layer sizes of (32, 8), ReLU activation, Adam optimizer, and maximum number of 100 iteration, and random state of 42. 
\end{itemize}

\paragraph{Detailed results: healthcare case study.}
We select a 3-month update frequency, and conduct this experiment for three classes of models: logistic regression, CART decision trees, and XGBoost. 
For each model type and each sequence, we graph the AUC and intra-sequence pairwise distance of the models comprising the slowly varying and greedy sequences in Fig. \ref{fig:intra_figure}. 
Differences in test AUCs of the SVML and greedy sequences grow smaller as we reduce accuracy tolerance. Indeed, for the $\alpha = 0.01$ SVML sequences, we observe only a 0.5\% reduction in AUC for logistic regression and negligible reductions for the other model types. However, we note an obvious advantage in stability measured by pairwise distance, ranging from 46\% (logistic regression) to 23\% (XGBoost). 


\begin{figure}[hbt]
\centering
\includegraphics[width=\linewidth]{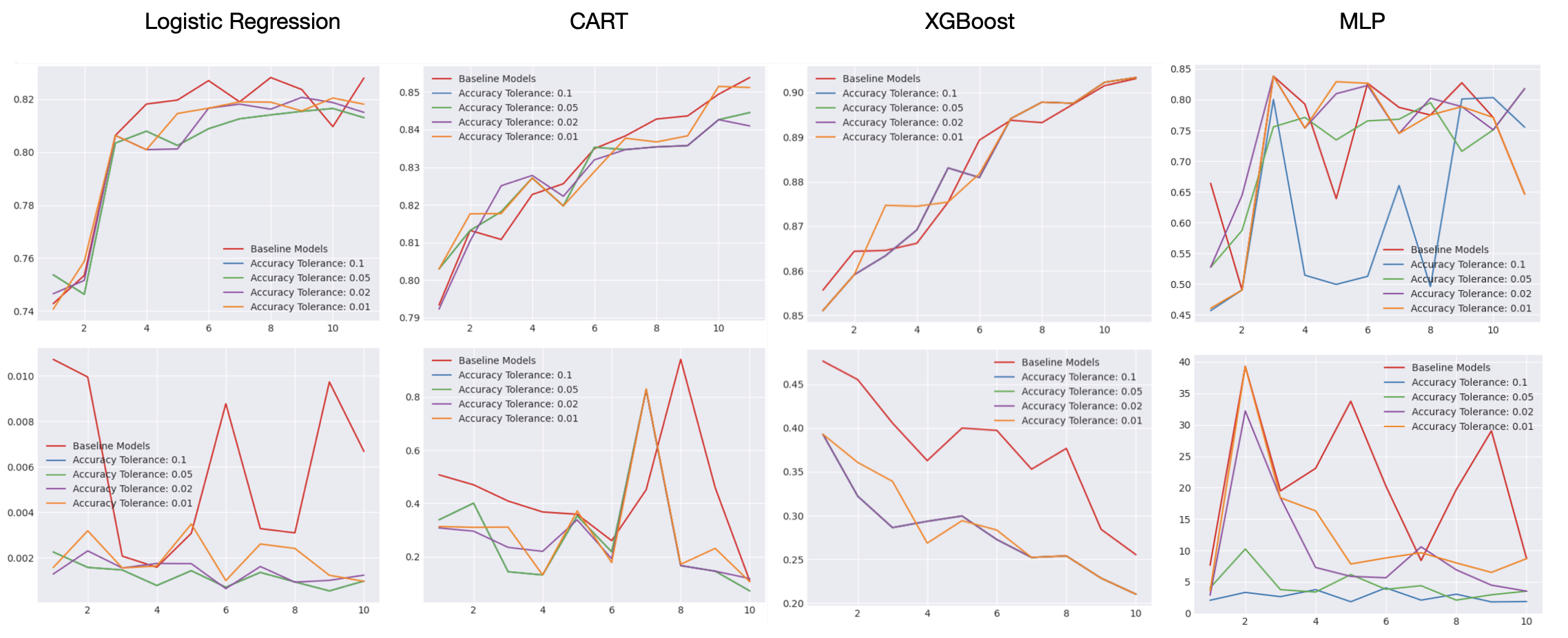}
    \caption{\small Healthcare case study: Test AUC (top) and distance between adjacent models (bottom) for various model/distance types}
    \label{fig:intra_figure}
\end{figure}


The intra-sequence stability results demonstrate the efficacy of SVML in minimizing the pairwise distance of adjacent models while sacrificing small model accuracy. E.g., by setting an accuracy tolerance of $0.01$, we guarantee that the validation AUC of chosen slowly varying models is within $1\%$ of the maximum: a bound that generalizes well experimentally to the test AUC. As expected, the AUC of the models and the pairwise distances for both SVML and greedy sequences tend to, respectively, increase and decrease over time as they are given more data to be trained on.

\paragraph{Robustness check: update frequency.} We empirically investigate the sensitivity of SVML (XGBoost) to different update frequencies. Fig. \ref{fig:distance-td} presents the results.
\begin{figure}[hbt]
\centering
\includegraphics[width=\linewidth]{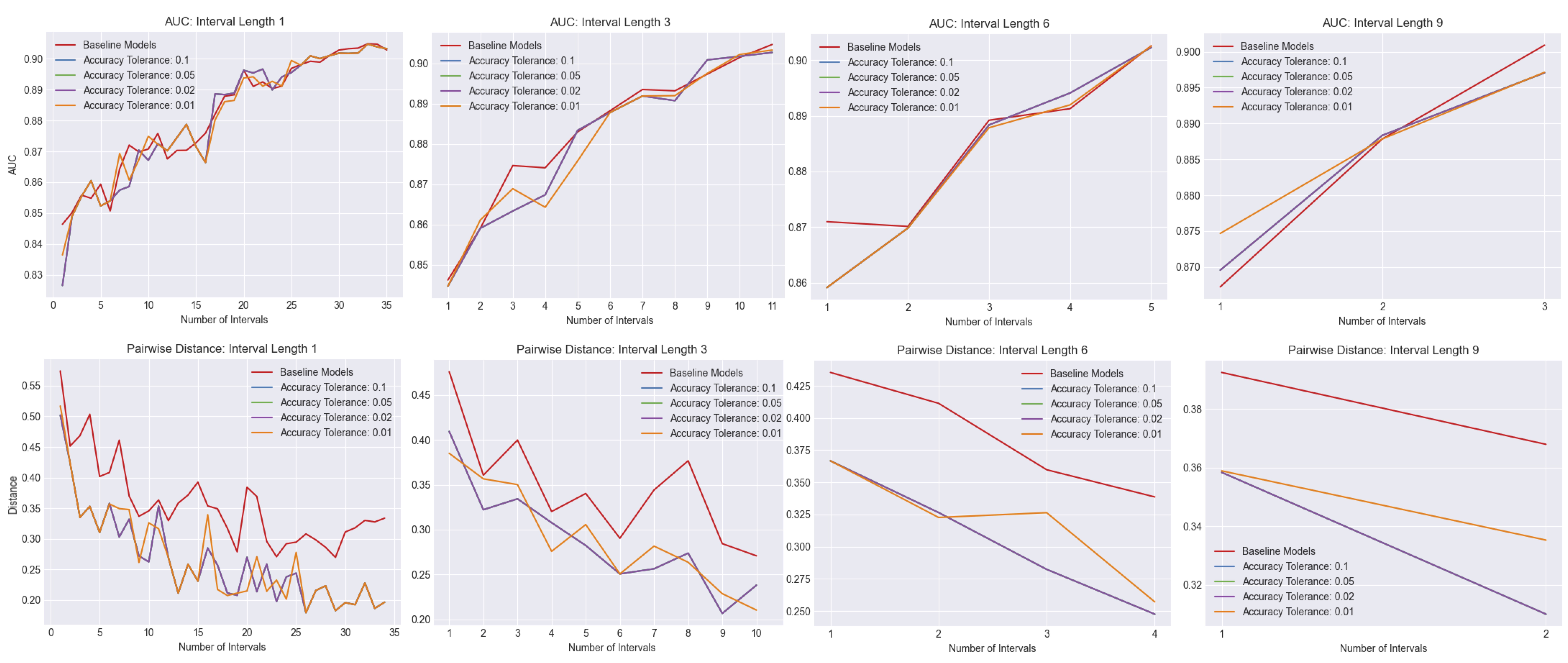}
\caption{\small Healthcare case study robustness check: AUC (top) and distance (bottom) between adjacent XGBoost models for various update frequencies.}
    \label{fig:distance-td}
\end{figure}

\paragraph{Robustness check: distance measure.} We investigate how well the analytical insights of an SVML sequence optimized for one distance measure (XGBoost with Gain importance) generalize to a different measure (SHAP) in terms of their stability across retraining iterations. Figure \ref{fig:inter_SHAP_evolv} presents the empirical distribution of the absolute deviation in features' SHAP values between adjacent models for both greedy and SVML sequences. The SVML approach demonstrates greater stability both iteration-to-iteration and across features.  Quantitatively, the mean stability loss for the greedy method is 0.0858 (std: 0.1013, max: 0.6246), whereas the SVML method achieves a lower mean of 0.0809 (std: 0.0937, max: 0.6060), confirming reduced fluctuations and improved consistency.
\begin{figure}[hbt]
\centering
\includegraphics[width=0.5\linewidth]{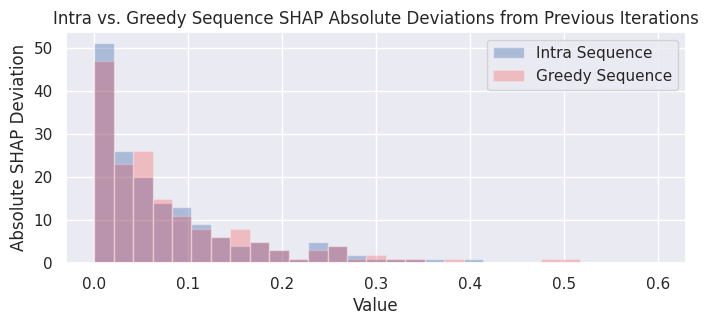}
    \caption{\small Healthcare case study robustness check: Empirical distribution of absolute deviation in SHAP between adjacent models.}
     \label{fig:inter_SHAP_evolv}
\end{figure}

\paragraph{Detailed results: vision case study.}
Fig. \ref{fig:yearbook_all} represents the result of the tradeoffs between AUC and distances across different time periods and accuracy thresholds for all four models. Similar to what we have observed for the healthcare dataset, there is a consistent trend of which models become more stable and accurate as we obtain more data. We observe high improvements using the SVML approach over greedy heuristic approaches across different models and different time periods with moderate sacrifice on model AUC performance. We also acknowledge the limitation that the for example in the CART case of period 9, it is potentially possible to see outlier performances due to the randomness of the dataset. However, the general trend remains similar to what we have observed previously. 

\begin{figure}[hbt]
\centering
\includegraphics[width=\linewidth]{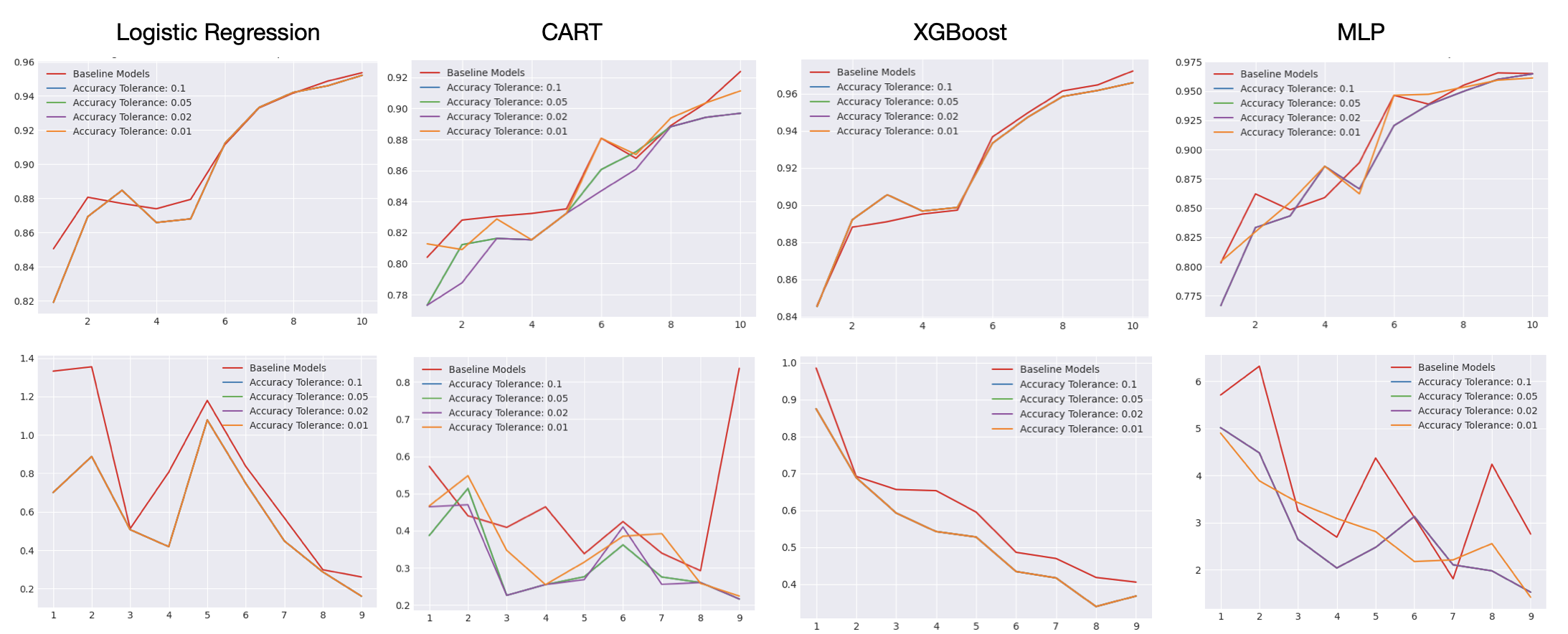}
    \caption{\small Vision case study: Test AUC (top) and distance between adjacent models (bottom) for various model/distance types}
    \label{fig:yearbook_all}
\end{figure}

\paragraph{Detailed results: language case study.}
We observe similar trends from previous studies, where we also see that the accuracy threshold does not have an as strong impact on the chosen models at the end. 

\begin{figure}[hbt]
\centering
\includegraphics[width=\linewidth]{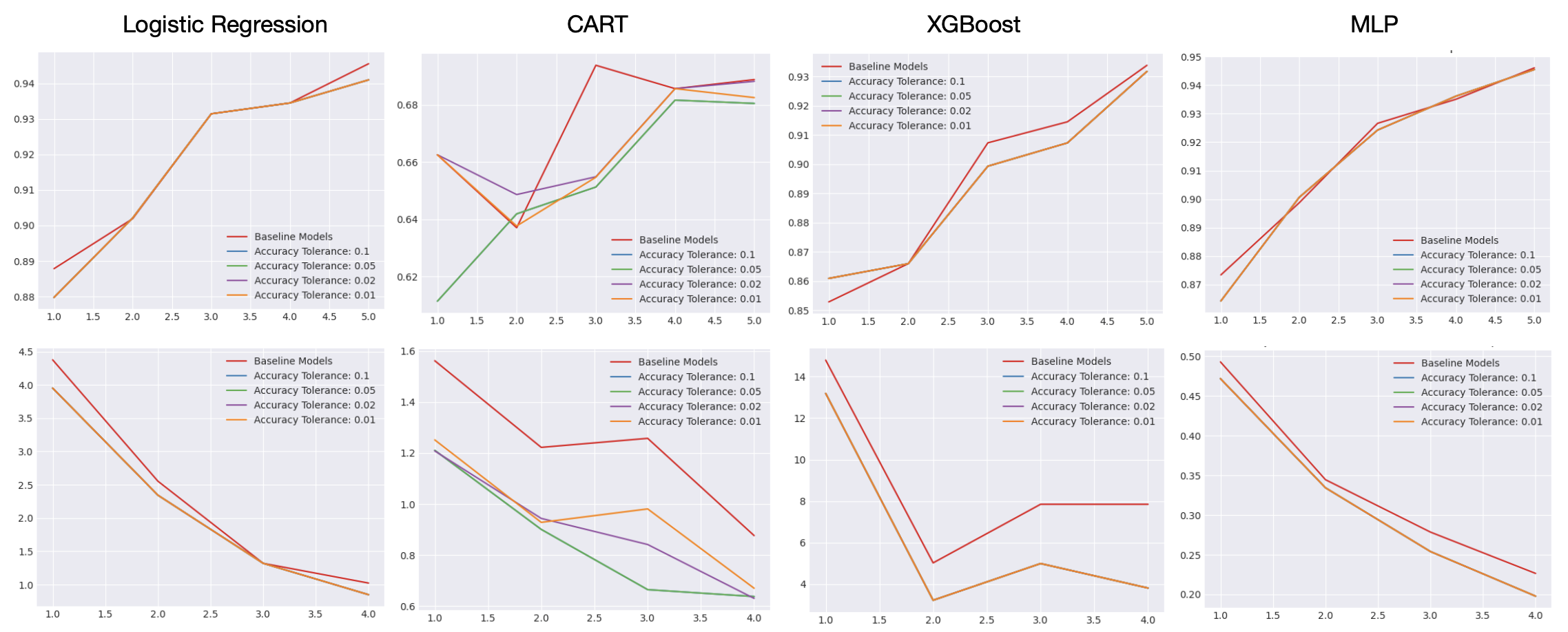}
    \caption{\small Language case study: Test AUC (top) and distance between adjacent models (bottom) for various model/distance types}
    \label{fig:lan_all}
\end{figure}

\end{document}